\documentclass{article}
\usepackage{amssymb}
\usepackage{ragged2e} 
\usepackage[hang,flushmargin]{footmisc} 
\usepackage{makecell}
\usepackage[protrusion=true,expansion=false]{microtype} 
\usepackage{graphicx}
\usepackage{multirow, multicol}
\usepackage{url}
\usepackage{caption}
\usepackage{float}
\usepackage{booktabs, tablefootnote}
\usepackage{footnote}
\makesavenoteenv{figure*}

\usepackage[table,xcdraw]{xcolor}
\usepackage{amsthm, amsmath,amssymb,amsfonts}
\usepackage{mathtools} 
\usepackage{mathrsfs}
\usepackage{algorithm}
\usepackage{algpseudocode}  
\usepackage{algorithmicx}

\usepackage{textcomp}
\usepackage{pifont}
\usepackage[utf8]{inputenc}
\usepackage{setspace,lineno}
\usepackage{listings}
\usepackage{soul}            
\sethlcolor{yellow}          
\usepackage[title]{appendix}
\usepackage[square,numbers,sort&compress]{natbib}
\usepackage{manyfoot}
\usepackage{xcolor}
\definecolor{customgreen}{HTML}{00B050}
\definecolor{captionblue}{HTML}{0070C0}  

\newcommand{\x}[1]{\textcolor{black}{#1}}

\DeclareCaptionLabelFormat{boldblue}{\textbf{\textcolor{captionblue}{#1 #2}}}
\usepackage{titlesec}
\titleformat{\section}
  {\normalfont\Large\bfseries\color{captionblue}}{\thesection}{1em}{}
\titleformat{\subsection}
  {\normalfont\large\bfseries\color{captionblue}}{\thesubsection}{1em}{}
\titleformat{\subsubsection}
  {\normalfont\normalsize\bfseries\color{captionblue}}{\thesubsubsection}{1em}{}

\usepackage{geometry}
\usepackage{amsmath} 
\allowdisplaybreaks[4]
\usepackage[protrusion=true,expansion=false]{microtype}
\usepackage{hyperref}
\hypersetup{
    colorlinks=true,
    linkcolor=blue,
    citecolor=blue,
    urlcolor=blue
}

\usepackage{natbib}
\title{A Physics-Informed, Behavior-Aware Digital Twin for Robust Multimodal Forecasting of Core Body Temperature in Precision Livestock Farming}
\author{
Riasad Alvi\textsuperscript{1,2}, 
Mohaimenul Azam Khan Raiaan\textsuperscript{3,a,*}, 
Sadia Sultana Chowa\textsuperscript{1,4}, \\
Arefin Ittesafun Abian\textsuperscript{1,2,a}, 
Reem E Mohamed\textsuperscript{5},  
Md Rafiqul Islam\textsuperscript{6}\\
Yakub Sebastian\textsuperscript{6},
Sheikh Izzal Azid\textsuperscript{7},
Sami Azam\textsuperscript{8,a,*}\\
\small
\textsuperscript{1}Applied Artificial INtelligence and Intelligent Systems (AAIINS) Laboratory, Dhaka 1217, Bangladesh \\
\small
\textsuperscript{2}Department of Computer Science and Engineering, United International University, Dhaka 1212, Bangladesh \\
\small
\textsuperscript{3}Department of Data Science and Artificial Intelligence, Monash University, Melbourne 3800, Australia \\
\small
\textsuperscript{4}Department of Software Systems \& Cybersecurity, Monash University, Melbourne 3800, Australia \\
\small
\textsuperscript{5}Energy and Resources Institute, Faculty of Science and Technology, Charles Darwin University, NSW 2000, Australia\\
\small
\textsuperscript{6}Faculty of Science and Technology, Charles Darwin University, Casuarina, NT, 0810 Australia \\
\small
\textsuperscript{7}School of Engineering and Energy, Murdoch University, Murdoch, WA 6150, Australia \\
\small
\textsuperscript{8}Energy and Resources Institute, Faculty of Science and Technology, Charles Darwin University, Darwin, NT, 0810, Australia
\small 
}
\date{} 
\begin{document}
\justifying
\twocolumn[
\maketitle
\begin{abstract}
\noindent 

Precision livestock farming requires accurate and timely heat stress prediction to ensure animal welfare and optimize farm management. This study presents a physics-informed digital twin (DT) framework combined with an uncertainty-aware, expert-weighted stacked ensemble for multimodal \x{forecasting} of Core Body Temperature (CBT) in dairy cattle. Using the high-frequency, heterogeneous MmCows dataset, the DT integrates an ordinary differential equation (ODE)-based thermoregulation model that simulates metabolic heat production and dissipation, a Gaussian process for capturing cow-specific deviations, a Kalman filter for aligning predictions with real-time sensor data, and a behavioral Markov chain that models activity-state transitions under varying environmental conditions. The DT outputs key physiological indicators, such as predicted CBT, heat stress probability, and behavioral state distributions are fused with raw sensor data and enriched through multi-scale temporal analysis and cross-modal feature engineering to form a comprehensive feature set. The predictive methodology is designed in a three-stage stacked ensemble, where stage 1 trains modality-specific LightGBM 'expert' models on distinct feature groups, stage 2 collects their predictions as meta-features, and at stage 3 Optuna-tuned LightGBM meta-model yields the final CBT \x{forecast}. Predictive uncertainty is quantified via bootstrapping and validated using Prediction Interval Coverage Probability (PICP). Ablation analysis confirms that incorporating DT-derived features and multimodal fusion substantially enhances performance. The proposed framework achieves a cross-validated $R^{2}$ of 0.783, F1 score of 84.25\% and PICP of 92.38\% for 2-hour ahead \x{forecasting}, providing a robust, uncertainty-aware, and physically principled system for early heat stress detection and precision livestock management.
\end{abstract}

\vspace{0.5em}
\vspace{1em}
]
{
\renewcommand{\thefootnote}{}
\footnotetext{%
\RaggedRight 
\textsuperscript{a}Equal Supervision\\
\textsuperscript{*}Correspondence: M. Raiaan (\href{mailto:mohaimenul.raiaan@monash.edu}{mohaimenul.raiaan@monash.edu}), \\ S. Azam (\href{mailto:sami.azam@cdu.edu.au}{sami.azam@cdu.edu.au}), 

}
}

\section{Introduction}
Dairy cattle experience heat stress when accumulated body heat exceeds their thermal regulation capacity, disrupting normal physiological stability \cite{shephard2023review, polsky2017invited}. This state reduces metabolic rates, and young animals face the most significant risk of adverse outcomes or death \cite{32, herbut2021effects}. Cattle may encounter heat stress in any climatic region, with its intensity varying throughout the year \cite{carvajal2021increasing, hasan2026impact}. Therefore, this has emerged as one of the most economically significant challenges facing modern agriculture, with projected global costs reaching \$30 billion by 2050 \cite{30}. Beyond immediate economic impacts, heat stress fundamentally compromises animal welfare, reduces milk yield by 174 ± 7 kg per cow per decade, compromises reproductive performance with 20–30\% drops in conception rates, and in extreme cases can be fatal \cite{chen2024effects,31}. When monitoring this condition, core body temperature (CBT) is the most reliable physiological indicator of heat stress. Effective CBT prediction enables proactive interventions before stress impacts milk yield, reproduction, or animal health \cite{ singaravadivelan2025navigating}. However, its accurate prediction from multi-modal sensor data remains a considerable challenge requiring urgent technological innovation to safeguard animal welfare and global food security \cite{shu2021recent, gomes2025review}.

Recent years have seen rapid advances in sensor technology, computational modeling, and machine learning (ML) that have significantly improved our capacity to detect and manage heat stress in livestock. State-of-the-art (SOTA) approaches include sensor-based monitoring systems that track environmental, physiological, and behavioral signals \cite{tangorra2024internet,17}, ML models that forecast panting duration or respiration rate based on meteorological inputs \cite{17}. Non-contact thermography, wearable accelerometers, and camera-based systems have also enabled automated, non-invasive detection of heat stress and behavior with high accuracy \cite{michelena2025review}. Advanced thermal imaging integrated with artificial intelligence has shown promise for non-invasive heat stress detection, with thermal signature methods providing more robust assessments than traditional discrete measurements \cite{pacheco2020thermal,pereira2024predictive}. 

Despite these achievements, significant limitations prevent existing methods from being widely adopted in operational dairy environments. Most critically, current ML approaches are predominantly data-centric and lack physiological grounding, rendering them vulnerable to sensor noise, equipment failures, and the ubiquitous problem of missing modality data that commonly occurs due to network instability, environmental contamination, and sensor degradation \cite{sharma2020machine,modak2025internet}. Existing models typically ignore the complex temporal dynamics and delayed effects of environmental conditions on individual animal physiology, missing critical windows where preventive interventions could be most effective \cite{mahmud2021systematic}. Another limitation is the absence of robust uncertainty quantification, which undermines confidence in predictions and restricts their utility for high-stakes decision-making in livestock management \cite{30}. The heterogeneous and multichannel nature of modern farm data also remains poorly integrated, with most systems failing to fuse information from diverse sensor modalities while maintaining interpretability effectively, a crucial requirement for adoption by farm managers who need to understand and trust automated recommendations \cite{oliveira2025heat}.

To address these limitations, we present a physics-informed DT framework that integrates first-principles physiological modeling with advanced ML architectures. Using high-frequency, heterogeneous sensor data from the MmCows dataset including vaginal CBT loggers, inertial measurement units (IMU), ultra-wideband (UWB) positioning, ankle accelerometers, environmental monitors, and milk yield records the framework aligns all streams to a uniform 1-minute resolution and constructs a hybrid DT comprising four tightly coupled components: (i)~an ODE-based thermoregulation model that captures metabolic heat production, activity-specific heat generation, and environmental dissipation; (ii)~a Gaussian process for cow-specific physiological deviations; (iii)~a Kalman filter that continuously synchronizes the DT with real sensor readings; and (iv)~a behavioral Markov chain that models activity-state transitions under varying environmental conditions. The DT-generated features predicted CBT, heat stress probability, and behavioral state probabilities are merged with raw sensor signals and enriched through multi-scale temporal statistics and cross-modal interaction features. These combined features feed a three-stage stacked ensemble: modality-specific LightGBM models produce per-group predictions (Stage~1), which serve as meta-features (Stage~2) for an Optuna-tuned LightGBM meta-model that yields the final 2-hour-ahead CBT \x{forecast} (Stage~3). A performance-weighted attention mechanism adaptively emphasizes the most informative modalities, enhancing resilience to incomplete sensor inputs. Uncertainty is quantified via bootstrapping and evaluated through Prediction Interval Coverage Probability (PICP). Experimental results on the MmCows dataset yield a cross-validated $R^{2}$ of 0.783, an F1 score of 84.25\%, and a PICP of 92.38\%, with ablation analyses confirming the contribution of both DT-derived features and multimodal fusion.

The major contributions of this paper are as follows:
\begin{itemize}
    \item Development of a physics-informed DT that integrates first-principles ODE models of cattle thermoregulation with Gaussian processes and Kalman filtering for real-time, personalized heat stress \x{forecasting}.
    \item Integration of multimodal sensor data into a principled, physiology-informed feature engineering pipeline that generates interpretable and temporally enriched features.
    \item Introduction of an uncertainty-aware, modality expert performance-weighted fusion mechanism within a stacked ensemble framework that adaptively prioritizes informative modalities and quantifies forecast uncertainty.
    \item Implementation of individualized prediction profiles that learn cow-specific parameters for personalized heat stress \x{forecasting} beyond one-size-fits-all approaches.
    \item Establishment of a rigorous evaluation framework using GroupKFold cross-validation with statistical significance testing to validate component contributions and prevent data leakage.
\end{itemize}

The remainder of this paper is organized as follows: Section \ref{sec:related_works} reviews related works in heat stress prediction and multimodal monitoring. Section \ref{sec:methodology} details the proposed physics-informed DT framework and the stacked ensemble methodology. Section \ref{sec:experimental_setup} describes the experimental setup, including the MmCows dataset and evaluation metrics. Section \ref{sec:results} presents the quantitative results and ablation studies. Section \ref{sec:discussion} discusses the findings and their implications for precision livestock farming, and Section \ref{sec:conclusion} concludes the paper with directions for future research.

\section{Related Works}
\label{sec:related_works}

In this section, we review recent studies on heat stress monitoring and CBT forecasting in dairy cattle, with a particular emphasis on data-centric prediction models, multimodal deep learning and behavioral monitoring approaches, and emerging digital-twin-based physiological frameworks. These techniques represent the principal methodological directions in Precision Livestock Farming (PLF) for early heat stress detection, where CBT elevation often precedes reductions in productivity and immune function \cite{sharma2020machine,17}.

\subsection{Data-Centric CBT Prediction and Environmental Index Models}
Data-centric CBT prediction and environmental index models form the foundation of heat stress monitoring in PLF by leveraging statistical and ML techniques to forecast thermal indicators from sensor data, as highlighted in recent work \cite{8,25,16,19,11,13,daniels2025detecting}. Li et al.~\cite{8} predicted CBT using a GWO--XGBoost ensemble, achieving an $R^{2}$ of approximately 0.54, a MAE of 0.232$^{\circ}$C, and an RMSE of 0.294$^{\circ}$C. In another paper, Lee et al.~\cite{25} proposed a multi-autoencoder framework for heat detection in dairy cows, aligning latent spaces across sensor types and combining anomaly detection with weakly supervised classification. This approach improved detection by ${\sim}$46\% over independent autoencoders and achieved an average $F1$ score of 70\%. Similarly, Chapman et al.~\cite{16} developed 24-hour heat stress forecasting models for feedlot cattle using accelerometer data, where an LSTM-based model outperformed logistic regression with lower MAE (${\sim}$0.08--0.15) and higher $R^{2}$ for panting duration prediction. Furthermore, Becker et al.~\cite{19} developed a heat stress scoring system using logistic regression, Gaussian na\"{i}ve Bayes, and random forest classifiers, with random forest achieving the best results at 88.8\% accuracy and an AUC of 94.5\%.

On the environmental index front, Georgiades et al.~\cite{11} produced hourly THI reconstructions via XGBoost trained on ERA5 reanalysis to better represent diurnal thermal load. Woodward et al.~\cite{13} matched rumen temperature events to weather data and found that simple regression indices produced many false positives (precision 9--27\%), while a Cubist model improved detection to 79\% sensitivity and 52\% precision. Finally, Daniels et al.~\cite{daniels2025detecting} identified a THI threshold of 74 for heat stress in peripubertal dairy heifers, with pasture housing producing abrupt, non-linear increases in respiration rate and body temperature at $\text{THI} = 74$ ($R^{2}$ of 0.704 and 0.578). However, reliance on THI alone and simple regression limits individualized CBT forecasting under real-world sensor variability. Despite promising accuracy, these data-centric models generally lack physiological interpretability and robustness to sensor dropout, limiting their deployment in operational PLF environments.

\subsection{Multimodal Deep Learning and Behavioral Monitoring Approaches}
Multimodal ML and deep learning approaches have demonstrated high classification accuracy by integrating complementary data streams, while behavioral monitoring provides additional proxies for thermal load, as highlighted in recent work \cite{2,3,4,5,6,7,12}. To begin with, Dhaliwal et al.~\cite{2} fused facial images (DenseNet-121) and accelerometer time series (LSTM) via multi-head attention, reporting 99.55\% accuracy for lameness detection; however, the study involved only six Holstein cows monitored for 21 days, which restricts external validity. In another paper, Tong et al.~\cite{3} employed adaptive fuzzy weighting within a multimodal evaluation framework, achieving validation accuracies above 90\% across environmental, feeding, and behavioral assessment modules. Similarly, Negreiro et al.~\cite{4} used YOLOv3 variants for calf posture and location classification, achieving recall often above 94\%. Furthermore, Sousa et al.~\cite{5} derived a thermal signature from infrared thermography, achieving 94.1\% accuracy with random forests for two-level heat stress classification in a controlled chamber experiment. While these results illustrate the potential of multimodal fusion, the high accuracies are predominantly from classification on small, controlled datasets and should not be conflated with continuous long-horizon CBT forecasting under field conditions. Moreover, these models typically assume synchronized, clean inputs, rarely addressing missing modalities or uncertainty propagation.

On the behavioral monitoring front, Eckhardt et al.~\cite{6} demonstrated moderate-to-strong climate--behavior correlations, including shade-seeking ($r = 0.66$) and nocturnal grazing shifts ($r = 0.64$) across genotypes. Szalai et al.~\cite{7} showed that reticulorumen and vaginal temperature, combined with rumination and activity patterns, can discriminate health states during the peripartum period but require individual calibration for accurate CBT forecasting. Finally, D\v{z}ermeikait\.{e} et al.~\cite{12} combined THI, milk temperature, rumination, and milk composition into composite heat-stress indicators (AUC $>$ 0.90) on ${\sim}$36,000 daily records from ${\sim}$200 cows, though daily resolution remains coarse for continuous forecasting. While these studies demonstrate promising capabilities, their reliance on limited datasets from single farms or specific breeds raises concerns regarding generalizability. Similar constraints apply to the MmCows dataset used in this study, which represents Holstein cows from a single geographic region.

\begin{table*}[ht!]
\centering
\caption{Comparison of the Proposed Framework with State-of-the-Art Methods.}
\label{tab:comparison}
\scriptsize
\definecolor{darkgreen}{rgb}{0.0, 0.5, 0.0}
\definecolor{darkred}{rgb}{0.7, 0.0, 0.0}
\begin{tabular}{l c c c c c}
\toprule
\textbf{Study} & \makecell{\textbf{Multimodal}\\\textbf{Fusion}} & \makecell{\textbf{Physics-}\\\textbf{Informed}} & \makecell{\textbf{Continuous CBT}\\\textbf{Forecasting}} & \makecell{\textbf{Uncertainty}\\\textbf{Quantification}} & \textbf{Personalization} \\
\midrule
Becker et al.\ \cite{19}       & \textcolor{darkgreen}{\ding{51}} & \textcolor{darkred}{\ding{55}} & \textcolor{darkred}{\ding{55}} & \textcolor{darkred}{\ding{55}} & \textcolor{darkred}{\ding{55}} \\
Georgiades et al.\ \cite{11}   & \textcolor{darkred}{\ding{55}}   & \textcolor{darkred}{\ding{55}} & \textcolor{darkred}{\ding{55}} & \textcolor{darkred}{\ding{55}} & \textcolor{darkred}{\ding{55}} \\
Woodward et al.\ \cite{13}  & \textcolor{darkred}{\ding{55}}   & \textcolor{darkred}{\ding{55}} & \textcolor{darkred}{\ding{55}} & \textcolor{darkred}{\ding{55}} & \textcolor{darkred}{\ding{55}} \\
Daniels et al.\ \cite{daniels2025detecting}      & \textcolor{darkred}{\ding{55}}   & \textcolor{darkred}{\ding{55}} & \textcolor{darkred}{\ding{55}} & \textcolor{darkred}{\ding{55}} & \textcolor{darkred}{\ding{55}} \\
Li et al.\ \cite{8}               & \textcolor{darkgreen}{\ding{51}} & \textcolor{darkred}{\ding{55}} & \textcolor{darkgreen}{\ding{51}} & \textcolor{darkred}{\ding{55}} & \textcolor{darkred}{\ding{55}} \\
Chapman et al.\ \cite{16}           & \textcolor{darkgreen}{\ding{51}} & \textcolor{darkred}{\ding{55}} & \textcolor{darkred}{\ding{55}} & \textcolor{darkred}{\ding{55}} & \textcolor{darkred}{\ding{55}} \\
Youssef et al. \cite{youssef2024}                       & \textcolor{darkgreen}{\ding{51}} & \textcolor{darkred}{\ding{55}} & \textcolor{darkred}{\ding{55}} & \textcolor{darkred}{\ding{55}} & \textcolor{darkgreen}{\ding{51}} \\
Zhang et al.\ \cite{zhang2025multimodal}         & \textcolor{darkgreen}{\ding{51}} & \textcolor{darkred}{\ding{55}} & \textcolor{darkred}{\ding{55}} & \textcolor{darkred}{\ding{55}} & \textcolor{darkred}{\ding{55}} \\
Kemp et al.\ \cite{kemp2025digital}              & \textcolor{darkgreen}{\ding{51}} & \textcolor{darkred}{\ding{55}} & \textcolor{darkred}{\ding{55}} & \textcolor{darkred}{\ding{55}} & \textcolor{darkgreen}{\ding{51}} \\
\midrule
\textbf{Proposed Framework} & \textcolor{darkgreen}{\ding{51}} & \textcolor{darkgreen}{\ding{51}} & \textcolor{darkgreen}{\ding{51}} & \textcolor{darkgreen}{\ding{51}} & \textcolor{darkgreen}{\ding{51}} \\
\bottomrule
\end{tabular}
\end{table*}

\subsection{Digital Twin-Based Physiological Frameworks}
Digital Twin (DT) frameworks have recently emerged as a promising direction for integrating real-time sensor data with physiological models in livestock research; however, existing implementations remain limited in scope, integration, and real-time feedback capabilities, as highlighted in recent work \cite{castillo2025bridging,youssef2024,zhang2025multimodal,kemp2025digital}. To begin with, Castillo et al.~\cite{castillo2025bridging} employed a DT for methane emission scenario-testing in dairy cows, but relied entirely on static Tier~2 regression models fitted to dietary inputs, with no real-time sensor fusion or physiological ODEs. In another paper, Youssef et al.~\cite{youssef2024} introduced the IUMENTA framework, a modular DT platform for estimating animal energy expenditure using software sensors and standard ML models (linear regression, random forests); however, it operates without mechanistic physiological constraints, Kalman-based state correction, or uncertainty quantification, and has not been applied to thermal stress forecasting. Similarly, Zhang et al.~\cite{zhang2025multimodal} addressed modality loss in multimodal dairy cow DTs through a cross-modal completion network for behavior classification, yet their framework is limited to behavioral state recognition and does not model the underlying thermal physiology or generate continuous CBT forecasts. Furthermore, Kemp et al.~\cite{kemp2025digital} constructed a DT of tropical dairy cattle that visualizes heat-stress impacts on feeding via per-animal linear regressions, providing correlational insights but no predictive forecasting capability, no mechanistic coupling between behavior and thermoregulation, and no calibrated uncertainty estimates. A recurring limitation across these and other multimodal livestock monitoring studies is the inadequate treatment of missing data, asynchronous sampling, and sensor noise; many approaches assume fully observed inputs or rely on simple forward-filling strategies, introducing bias in continuous physiological forecasting.

Our proposed framework addresses several key limitations observed in existing studies by offering novel architectural solutions, as presented in Table~\ref{tab:comparison}. Unlike all reviewed approaches, our framework embeds physiological grounding through a thermodynamic ODE rooted in the First Law of Thermodynamics, ensuring that predictions respect energy-conservation constraints rather than relying solely on statistical correlations. Moreover, the ODE--Kalman filter loop produces continuous, minute-resolution CBT trajectories over a two-hour forecast horizon, moving beyond the discrete classification paradigm that dominates the current literature. A Gaussian process quantifies residual epistemic uncertainty and bootstrap-based predictive intervals yield a PICP of 92.38\%, providing calibrated confidence bounds that no reviewed method offers. Furthermore, a performance-weighted mechanism assigns modality-specific reliability scores derived from held-out validation error, replacing ad-hoc feature concatenation with principled, adaptive fusion. Within this framework, a behavioral Markov model dynamically modulates the thermal ODE according to each animal's activity-state transitions, enabling individualized forecasts that account for inter-animal variability in thermoregulatory response.
 
While data-centric models excel at extracting statistical patterns from sensor data and multimodal deep learning approaches enhance classification accuracy through feature fusion, both categories share critical limitations: they lack physiological grounding, do not support continuous long-horizon CBT forecasting, omit uncertainty quantification, employ heuristic rather than reliability-weighted fusion strategies, and neglect personalization despite substantial inter-animal variability. Similarly, existing DT frameworks, while promising in concept, have not incorporated physics-based thermoregulation models, delivered continuous CBT forecasts, or provided calibrated uncertainty quantification. These gaps directly motivate the development of our proposed physics-informed DT pipeline, which introduces thermodynamic modeling, adaptive fusion, and uncertainty-aware prediction modules to comprehensively address these shortcomings.

\section{Methodology}
\label{sec:methodology}

This section describes the complete multimodal pipeline for physiological monitoring and heat-stress prediction in dairy cattle. 

\begin{figure*}[!ht]
    \centering
    \includegraphics[width=0.95\textwidth]{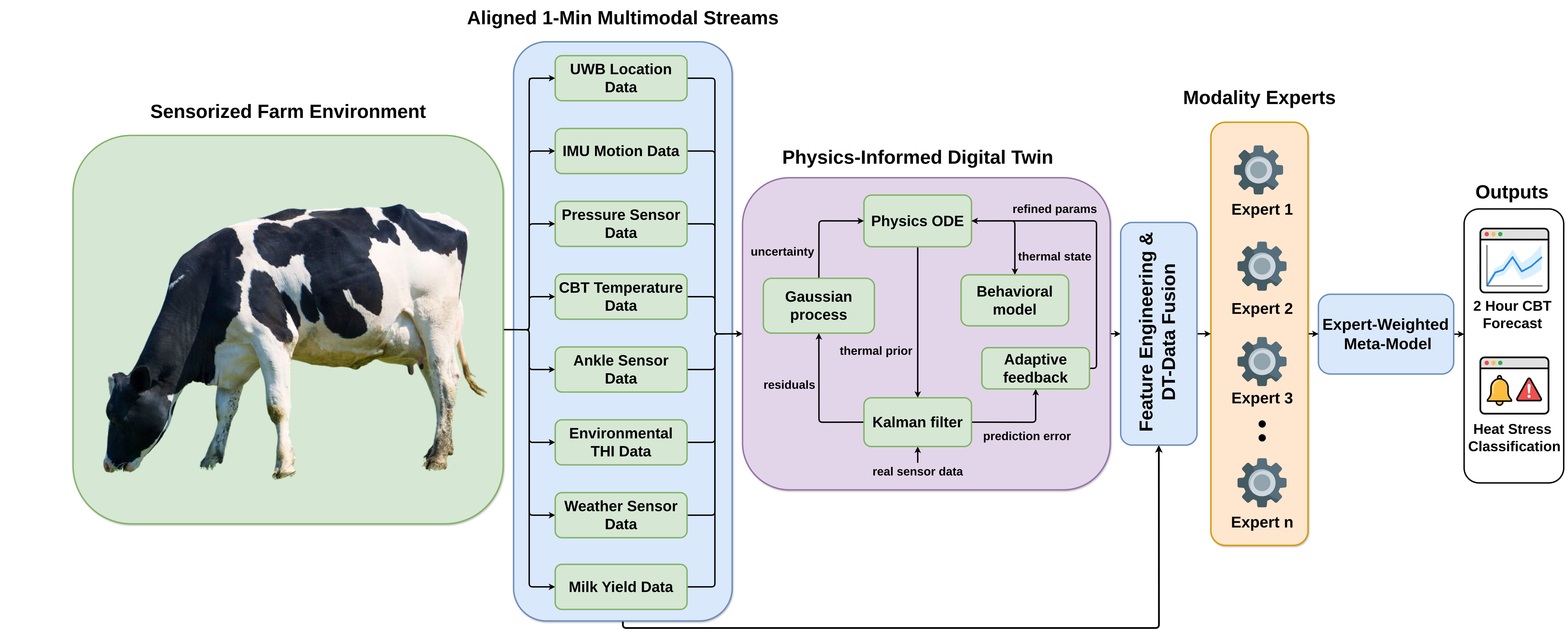}
    \caption[High-level architecture of the proposed multimodal forecasting framework]{High-level architecture of the proposed multimodal forecasting framework. Raw data from cow-worn and environmental sensors undergo temporal alignment and preprocessing before being fed into a Physics-Informed Digital Twin. The digital twin dynamically couples a behavioral Markov model, a thermodynamic ODE, a Kalman filter, and a Gaussian process to generate uncertainty-aware physiological and behavioral state features. These are fused with multi-scale engineered features to train modality-specific experts and an optimized meta-model, ultimately yielding continuous 2-hour ahead core body temperature (CBT) forecasts and heat-stress classification. \textit{Copyright Free Cow Image source:} Freepik\protect\footnote{\url{https://www.freepik.com/free-photo/cow-grazing-green-meadow_11244779.htm}}.}
    \label{fig:methodology_architecture}
\end{figure*}

The proposed framework integrates multimodal sensor data collection, synchronized preprocessing, DT simulation, multi scale feature engineering, and an expert-driven ensemble meta-model with uncertainty quantification. An overview of this high level multimodal pipeline is illustrated in Figure~\ref{fig:methodology_architecture}. The overall end-to-end architecture, encompassing all stages of the framework, is illustrated in Figure ~\ref{fig:meth}.



\subsection{Multimodal Sensor Data Collection}

The foundation of the proposed framework is a heterogeneous sensing infrastructure designed to continuously monitor the physiological, behavioral, and environmental parameters of dairy cattle. Multiple sensor modalities operate concurrently to capture distinct yet complementary aspects of an individual cow’s state, thereby enabling cross-domain reasoning through multimodal data fusion \cite{6}. Each modality is represented as a multivariate time-series tensor , as defined in Eq.~\ref{eq:modality_tensors}:

\begin{align}
X_{\text{uwb}} &\in \mathbb{R}^{B \times T \times 3}, \nonumber\\
X_{\text{immu}} &\in \mathbb{R}^{B \times T \times 6}, \nonumber\\
X_{\text{pressure}} &\in \mathbb{R}^{B \times T \times 1}, \nonumber\\
X_{\text{cbt}} &\in \mathbb{R}^{B \times T \times 1}, \nonumber\\
X_{\text{ankle}} &\in \mathbb{R}^{B \times T \times 1}, \nonumber\\
X_{\text{thi}} &\in \mathbb{R}^{B \times T \times 1}, \nonumber\\
X_{\text{weather}} &\in \mathbb{R}^{B \times T \times 5}, \nonumber\\
X_{\text{milk}} &\in \mathbb{R}^{B \times T \times 1}.
\label{eq:modality_tensors}
\end{align}

In this formulation, $B$ denotes the number of monitored animals, and $T$ represents the total number of time steps after temporal alignment. The multimodal configuration encompasses eight complementary sensing systems:

\begin{figure*}[ht!]
\centering
\includegraphics[scale=0.315]{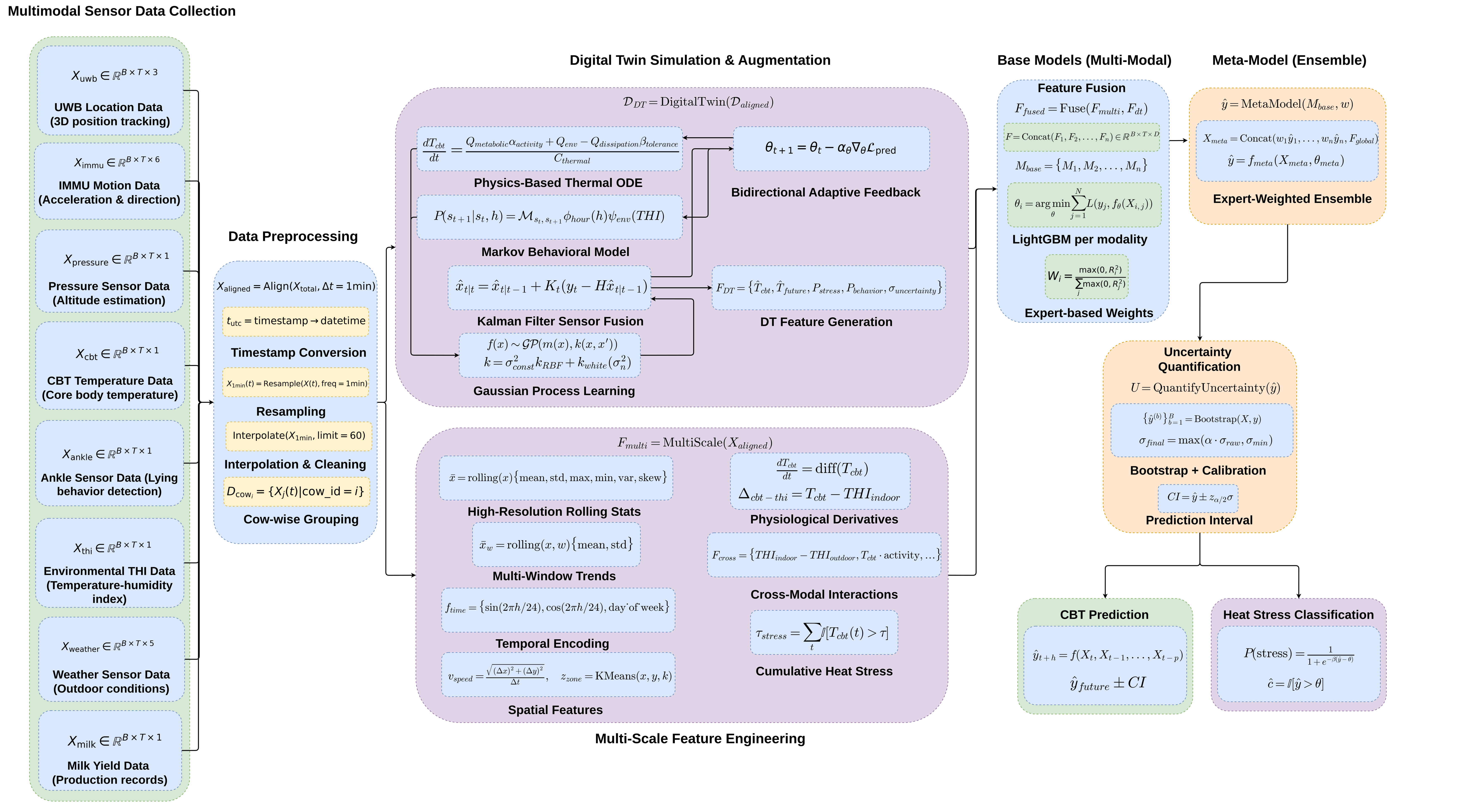}
\caption{Overview of the proposed pipeline. Sensor data from eight modalities are aligned and preprocessed, then passed through a physics-informed DT that models thermal and behavioral dynamics. Multi-scale features are engineered and fed into modality-specific base models. An expert-weighted ensemble fuses predictions, with bootstrap-based uncertainty quantification yielding final CBT forecasts and heat-stress risk estimates.}
\label{fig:meth}
\end{figure*}

(1) ultra-wideband (UWB) transponders for three-dimensional spatial localization $(x, y, z)$; (2) inertial measurement and magnetic units (IMMU) for capturing six degrees of freedom in linear acceleration and orientation; (3) barometric pressure sensors for altitude estimation and stall-level positioning; (4) vaginal or intravaginal temperature sensors measuring core body temperature (CBT) with sub-minute precision; (5) ankle-mounted tilt sensors for discriminating posture and activity transitions such as lying, standing, or walking; (6) temperature-humidity index (THI) sensors for indoor microclimatic monitoring; (7) environmental weather stations measuring outdoor meteorological variables including ambient temperature, humidity, wind speed, and solar radiation; and (8) automated milking systems recording daily milk yield and milking time as physiological productivity indicators.

Together, these multimodal data streams provide a comprehensive depiction of the animal’s thermoregulatory state within its behavioral and environmental context. By integrating high-frequency physical measurements with slow-varying production records, the system establishes a fine-grained basis for dynamic modeling of heat stress, behavioral adaptation, and metabolic response. This multimodal architecture not only enables data-level complementarity but also facilitates physics-informed DT modeling through cross-domain correlations among internal, behavioral, and external variables.

Although these modalities provide complementary views of the animal's thermal state, they inherently exhibit practical limitations that influence data integrity. Wearable devices, such as UWB transponders and pressure sensors, frequently experience intermittent signal dropout due to battery depletion, physical detachment, radio interference, or environmental occlusion. Furthermore, IMMU and ankle sensors are susceptible to calibration drift and transient outliers caused by physical impacts or device slippage. Even CBT loggers, despite providing high temporal fidelity, can introduce measurement noise and occasional missing data segments. To maintain methodological transparency and robust forecasting, our preprocessing and DT fusion pipelines are explicitly designed to account for these inconsistencies through short-gap interpolation, long-gap uncertainty propagation, and outlier-robust aggregation (Section \ref{subsec:preprocessing}-\ref{subsec:digitaltwin}).

\subsection{Data Preprocessing}
\label{subsec:preprocessing}

Comprehensive preprocessing of multimodal data is required to ensure integrity and temporal consistency across heterogeneous sensors. Our pipeline performs temporal alignment, timestamp normalization, resampling, interpolation, and identity-based grouping to transform asynchronous raw measurements into coherent, leak-free time-series sequences for digital-twin simulation and downstream learning tasks. Each operation is applied systematically to maintain temporal causality and to prevent cross-animal contamination in model training.

\subsubsection{Temporal Alignment and Resampling}

Sensor devices operate at different sampling frequencies ranging from milliseconds  to several minutes. To establish a unified temporal scale, all sensor signals are synchronized using a fixed sampling interval of $\Delta t = 1$ minute. The resulting temporally aligned dataset is defined in Eq.~\ref{eq:temporal_alignment_resampling}:
\begin{equation}
X_{\text{aligned}} = \text{Align}(X_{\text{total}}, \Delta t = 1\text{min})
\label{eq:temporal_alignment_resampling}
\end{equation}
This alignment ensures that multimodal observations correspond to the same temporal index, enabling reliable cross-modal feature construction and DT parameter updates at consistent time steps. 

\subsubsection{Timestamp Conversion}

All raw timestamps are standardized to ensure temporal consistency across sensor modalities. Each device-recorded timestamp is converted from its local clock reference into Coordinated Universal Time (UTC). This normalization guarantees globally comparable time indices across all data sources, as expressed in Eq.~\ref{eq:timestamp_conversion}:
\begin{equation}
t_{\text{utc}} = \text{timestamp} \rightarrow \text{datetime}
\label{eq:timestamp_conversion}
\end{equation}
This conversion reconciles time-zone offsets and synchronization drift among distributed acquisition systems. All subsequent temporal operations, including resampling and window-based computation, are performed in UTC to maintain chronological coherence across modalities.

\subsubsection{Resampling and Interpolation}

Following temporal alignment, all modalities with irregular or high-frequency sampling rates are resampled to a uniform 1-minute interval. This produces a consistent temporal structure across the dataset, as defined in Eq.~\ref{eq:resampling_operation}:

\begin{equation}
X_{\text{resampled}}(t) = \text{Resample}\!\left(X(t),\, \text{freq}=1\text{ min}\right)
\label{eq:resampling_operation}
\end{equation}

To maintain the causal integrity of the time-series forecasting, missing values were handled using linear interpolation strictly within short windows ($< 60$\,s), as shown in Eq.~\ref{eq:bounded_interpolation}:

\begin{equation}
\text{Interpolate}\!\left(X_{\text{immu}},\, \text{limit}=60\right)
\label{eq:bounded_interpolation}
\end{equation}

For longer gaps, no forward-filling was applied to prevent temporal leakage; instead, the Kalman Filter (Section \ref{subsec:kalman}) propagates the last known state variance forward, ensuring the model relies only on historical context and not future data points. During these extended gaps, timestamps are preserved, but measurement updates for the affected modality are conditionally skipped, allowing the DT and Kalman fusion process to rely on the underlying process model rather than synthetic observations. In parallel, modality-level missingness indicators (e.g., missing flags and time-since-last-observation signals) are explicitly exposed to the downstream learners so that unreliable inputs can be adaptively down-weighted during ensemble fusion. This comprehensive strategy distinguishes between transient signal loss and extended sensor dropouts, ensuring that prolonged missingness appropriately increases epistemic uncertainty and widens confidence intervals, all without introducing non-causal trends into the reconstructed time series.

\subsubsection{\x{Temperature--Humidity Index Computation}}

\x{The Temperature--Humidity Index is computed from the temporally aligned indoor temperature and relative-humidity streams at each time step as}
\begin{equation}
\x{\mathrm{THI} = 0.8\,T + \frac{\mathrm{RH}}{100}\,\bigl(T - 14.4\bigr) + 46.4,}
\label{eq:thi_definition}
\end{equation}
\x{where $T$ is the ambient air temperature ($^{\circ}$C) and $\mathrm{RH}$ the relative humidity (\%). Equation~\ref{eq:thi_definition} is the National Research Council temperature--humidity index~\cite{nrc1971,dikmen2009temperature}, as used for the indoor microclimate of the MmCows dataset~\cite{vu2024mmcows}. The index is computed for both the indoor and outdoor measurements, yielding $\mathrm{THI}_{\mathrm{indoor}}$ and $\mathrm{THI}_{\mathrm{outdoor}}$; the indoor value drives the thermodynamic ODE (Section~\ref{subsec:digitaltwin}) and the cross-modal differential of Eq.~\ref{eq:thi_temperature_differential}.}

\subsubsection{Cow-wise Grouping and Data Integrity}

To ensure identity consistency and to prevent inter-animal leakage during training, all temporally aligned multimodal sequences are grouped by cow identity , as defined in Eq.~\ref{eq:cow_grouping}:

\begin{equation}
D_{\text{cow}} = \{\, X_j(t) \mid \text{cow\_id} = i \,\}
\label{eq:cow_grouping}
\end{equation}

Each subset $D_{\text{cow}}$ forms a self-contained multimodal record of an individual animal’s time-series data. Group-level operations such as normalization, rolling window computation, and DT parameter updates are subsequently performed independently within each cow’s dataset.

\vspace{0.2em}
Through these preprocessing stages, the raw heterogeneous sensor signals are transformed into clean, temporally consistent, and structurally uniform data streams. This preprocessed dataset forms the analytical substrate for physics-informed DT modeling and multimodal feature extraction in subsequent stages of the framework.

\begin{figure*}[ht!]
\centering
\includegraphics[scale=0.2]{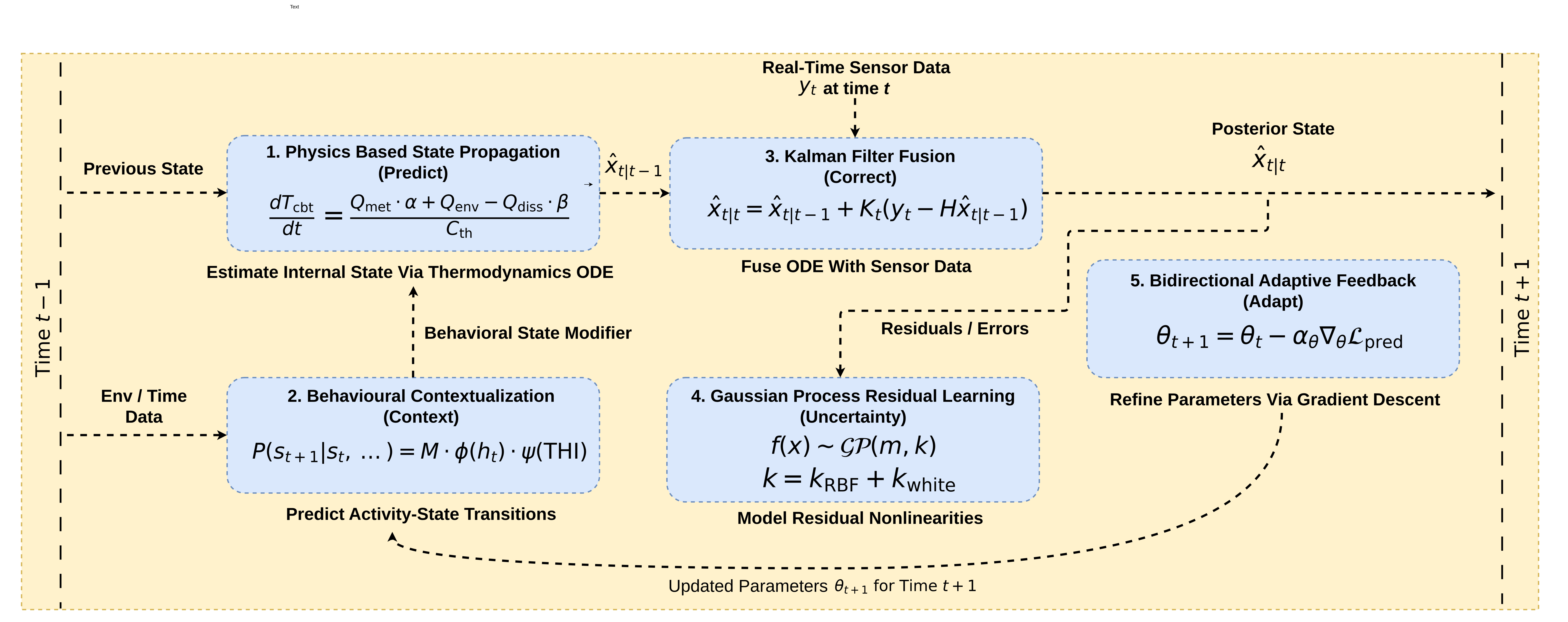}
\caption{Internal Computational Loop of the physics-informed DT. The diagram illustrates the recursive state estimation cycle performed at each time step $t$. It unifies the mechanistic prediction (ODE), behavioral context (Markov), sensor correction (Kalman Filter), and uncertainty quantification (Gaussian Process) into a closed-loop adaptive system.}
\label{fig:DT Loop}
\end{figure*}


\subsection{Digital Twin Simulation}
\label{subsec:digitaltwin}

The DT serves as the central computational core of the proposed multimodal framework. It functions as a dynamic, physics-informed, and data-driven virtual replica of each animal, integrating physical thermoregulation models, behavioral dynamics, probabilistic inference, and uncertainty-aware feature generation. By continuously coupling sensor-derived observations with internal physiological simulation, the DT provides a mechanistic bridge between environmental conditions, animal behavior, and thermal physiology \cite{purcell2023digital,arulmozhi2024reality}. This hybrid design allows real-time adaptation to varying conditions and supports predictive reasoning beyond the directly observed data. Figure~\ref{fig:DT Loop} illustrates the recursive simulation loop executed at each time step $t$. First, the Physics-Based ODE estimates the \textit{a priori} thermal state $\hat{x}_{t|t-1}$, modulated by dynamic activity probabilities from the Behavioral Markov Model. This prior is fused with real-time sensor data $y_t$ via the Kalman Filter to minimize estimation error and produce the posterior state $\hat{x}_{t|t}$. Residuals are mapped by the Gaussian Process to quantify uncertainty, while Bidirectional Feedback updates the physiological parameters $\theta$ via gradient descent to adapt to the animal's evolving state.

\subsubsection{Physics-Based Thermal ODE}

The evolution of a cow’s core body temperature (CBT) is modeled using a first-order energy balance formulation that accounts for metabolic heat generation, environmental thermal loading, and physiological heat dissipation \x{\cite{li2021mechanistic}}. The governing ordinary differential equation (ODE) is given in Eq.~\ref{eq:thermal_ode_physics}:

\begin{equation}
\frac{dT_{\text{cbt}}}{dt}
=
\frac{
Q_{\text{metabolic}}\, \alpha_{\text{activity}}
+ Q_{\text{env}}
- Q_{\text{dissipation}}\, \beta_{\text{tolerance}}
}{
C_{\text{thermal}}
}
\label{eq:thermal_ode_physics}
\end{equation}


In this formulation, $Q_{\text{metabolic}}$ denotes internal heat generation driven by basal metabolism and physical activity, scaled by an activity-dependent multiplier $\alpha_{\text{activity}}$ derived from IMMU acceleration patterns. The term $Q_{\text{env}}$ captures environmental heat load computed from the indoor temperature–humidity index (THI), which reflects ambient thermal stress. Heat dissipation is represented by $Q_{\text{dissipation}}$, modulated by the cow-specific thermotolerance coefficient $\beta_{\text{tolerance}}$, which encodes inter-animal variability in cooling efficiency. The dissipation term captures active physiological cooling (vasodilation, respiratory evaporation) and is modeled as $Q_{\text{dissipation}} = k_d \bigl( T_{\text{cbt}} - T_{\text{set}} \bigr)$, where $k_d$ is a lumped heat-loss coefficient and $T_{\text{set}}$ is the thermoneutral set-point for Holstein cattle. The denominator $C_{\text{thermal}}$ corresponds to the effective thermal capacity of the animal, representing the amount of energy required to change its internal temperature. To ensure physiological plausibility, the components are defined as follows:
$Q_{\text{metabolic}} = M_{\text{basal}} + \gamma \cdot A(t)$, where $M_{\text{basal}}$ is the species-specific basal
metabolic rate and $A(t)$ is the dynamic activity magnitude derived from the IMMU. The environmental load is modeled as $Q_{\text{env}} = k_{\text{th}} \bigl( T_{\text{eff}}(\mathrm{THI}) - T_{\text{cbt}} \bigr)$ .

Here, \( T_{\text{eff}} \) represents a linear mapping function that converts the unitless
Temperature--Humidity Index (THI) into an equivalent environmental temperature
(\(^\circ\mathrm{C}\)), ensuring dimensional consistency in the thermal gradient calculation. The heat capacity $C_{\text{thermal}}$ is initialized based
on the average weight of Holstein cows ($650\,\text{kg} \times 3.5\,\text{kJ}/\text{kg}^{\circ}\text{C}$) and fine-
tuned per animal \x{\cite{jan2025holstein}}. Parameters $\beta_{\text{tolerance}}$ and $\alpha_{\text{activity}}$ are learnable
coefficients \x{constrained to physiological ranges informed by} \cite{li2021mechanistic,zhou2022development,foroushani2022thermodynamic} to ensure the solver remains within biological
constraints. 

From a physiological and thermodynamic perspective, these learnable coefficients serve distinct roles in shaping the modeled heat dynamics. The activity multiplier $\alpha_{\text{activity}}$ modulates metabolic heat generation to reflect the increased heat production during locomotion or agitation, as inferred from IMMU-derived activity probabilities. Meanwhile, the thermotolerance coefficient $\beta_{\text{tolerance}}$ captures inter-animal variability in cooling efficiency (e.g., individual differences in heat dissipation capacity), thereby enabling personalization beyond a one-size-fits-all thermoregulation model. Furthermore, the effective heat capacity $C_{\text{thermal}}$ governs how rapidly the CBT changes in response to the net heat flux, allowing for bounded per-animal adaptation while strictly preserving physical plausibility. Together, these parameter constraints ensure that the learned dynamics remain biologically interpretable while supporting stable, causal online updates.

\x{The ODE is integrated with a fixed time step of $\Delta t = 1$~min, matching the resampling interval of the aligned data; its parameters and their initial values are listed in Table~\ref{tab:ode_params}.}

\begin{table}[ht]
\centering
\caption{Parameters of the physics-based thermal ODE (Eq.~\ref{eq:thermal_ode_physics}). Each parameter's status (fixed, calibrated, or learned per cow) and source are specified in the text.}
\label{tab:ode_params}
\scriptsize
\setlength{\tabcolsep}{4pt}
\renewcommand{\arraystretch}{1.15}
\begin{tabular}{lp{0.33\columnwidth}p{0.42\columnwidth}}
\toprule
\textbf{Parameter} & \textbf{Description} & \textbf{Initialisation} \\
\midrule
$C_{\text{thermal}}$
& Effective heat capacity
& $\approx 2275$ kJ\,${}^{\circ}$C$^{-1}$ ($650$ kg $\times$ $3.5$ kJ\,kg$^{-1}$\,${}^{\circ}$C$^{-1}$) \\
$T_{\text{set}}$
& Thermoneutral set-point
& $38.5\,^{\circ}$C \\
$M_{\text{basal}}$
& Basal metabolic heat
& $\approx 1240$ W (600 kg cow) \\
$\alpha_{\text{activity}}$
& Activity heat multiplier
& 1.0 (lying), 1.2 (standing), 1.4 (feeding), 1.8 (walking) \\
$\beta_{\text{tolerance}}$
& Thermotolerance scaling
& 1.0 \\
$k_{\text{th}}$
& Environmental coupling coefficient
& Calibrated \\
$k_d$
& Heat-loss coefficient
& Calibrated \\
\bottomrule
\end{tabular}
\end{table}

\x{The body-core specific heat, thermoneutral set-point, and basal metabolic heat in Table~\ref{tab:ode_params} follow mechanistic thermal-balance models of dairy cattle~\cite{li2021mechanistic,zhou2022development}, with a basal heat production of $\approx$1240~W for a 600~kg cow~\cite{li2021mechanistic}. The environmental coupling expresses the THI as an equivalent temperature, consistent with the apparent-temperature framework of~\cite{foroushani2022thermodynamic}, whereas $k_{\text{th}}$ and $k_d$ are calibrated to the cohort. All parameters except the personalized set $\{C_{\text{thermal}},\, \alpha_{\text{activity}},\, \beta_{\text{tolerance}}\}$ are shared across animals; these three are adapted for each animal through the online feedback rule (Eq.~\ref{eq:dt_feedback_update}) using only that animal's historical residuals, within the physiological ranges noted above.}





\subsubsection{Bidirectional Adaptive Feedback}

The DT operates in a closed-loop configuration in which its physiological parameters $\theta$ are continuously refined based on discrepancies between simulated and observed measurements \cite{zhang2025comprehensive}. This bidirectional feedback enables real-time personalization and model stability. The adaptive parameter update rule is given in Eq.~\ref{eq:dt_feedback_update}:

\begin{equation}
\theta_{t+1}
=
\theta_t
-
\alpha_\theta \nabla_\theta \mathcal{L}_{\text{prediction}}
\label{eq:dt_feedback_update}
\end{equation}

Here, $\alpha_\theta$ denotes the adaptive learning rate, and $\mathcal{L}_{\text{prediction}}$ quantifies the error between DT-predicted and sensor-observed core body temperature (CBT). By iteratively adjusting $\theta$ to reduce this prediction loss, the DT adapts its internal thermal and behavioral dynamics to reflect evolving metabolic conditions, environmental stressors, and individual animal variability. Parameter updates for the forecast at time $t$ are derived solely from errors observed at time $t - 1$ and earlier. No ground truth data from the forecast window $[t, t + h]$ is utilized for parameter adaptation. In order to ensure valid forecasting and prevent look-ahead bias, parameter updates during the test phase are performed in a strictly online manner; the model relies only on historical error gradients (computed at time $t$ or earlier) to refine the physiological state $\theta_t$ prior to generating the forecast for the future horizon $t + h$. This bidirectional learning mechanism ensures that the DT remains physiologically consistent and responsive during continuous monitoring.





\subsubsection{Markov Behavioral Model}

Animal behavior exerts a direct influence on heat exchange processes and must therefore be incorporated into the dynamics of the DT. To represent behavioral transitions, the DT integrates a first-order Markov model that probabilistically predicts activity-state changes under varying temporal and environmental contexts \cite{mor2021systematic}. The transition formulation is given in Eq.~\ref{eq:markov_behavior}:

\begin{equation}
P(s_{t+1} \mid s_t, h_t)
=
M_{s_t, s_{t+1}} \,
\phi_{\text{hour}}(h_t)
\, \psi_{\text{env}}(\text{THI})
\label{eq:markov_behavior}
\end{equation}

In this formulation, $s_t$ and $s_{t+1}$ denote consecutive behavioral states (lying, standing, walking, feeding). The term $M_{s_t, s_{t+1}}$ represents the baseline state-transition matrix estimated from historical behavioral sequences. The temporal factor $\phi_{\text{hour}}(h_t)$ captures diurnal activity dynamics, while the environmental modulation term $\psi_{\text{env}}(\text{THI})$ accounts for heat-stress–induced behavioral adjustments. The resulting product is row-wise re-normalized so that $\sum_{s_{t+1}} P(s_{t+1} \mid s_t, h_t) = 1$, preserving a valid probability distribution over successor states. Together, these components allow the DT to generate realistic, context-aware behavioral trajectories that co-evolve with the animal’s thermophysiological state.

\x{The behavioral states are derived from the cow-worn motion sensors (the ankle logger's lying/non-lying signal and the IMMU-derived activity magnitude), discretized into the four states above. The baseline transition matrix $M_{s_t,s_{t+1}}$ is estimated for each animal by maximum-likelihood counting of the observed state transitions with Laplace smoothing and row normalization, and the modulation factors $\phi_{\text{hour}}(h_t)$ and $\psi_{\text{env}}(\mathrm{THI})$ are estimated empirically from the same per-animal sequences. The behavioral component is thereby personalized to each cow, and its joint contribution with the remaining digital-twin components is reflected in the ablation of Table~\ref{tab:ablation_dt}.}





\subsubsection{Kalman Filter Sensor Fusion}
\label{subsec:kalman}
Sensor measurements are inherently noisy and may suffer from communication dropouts or gradual drift. To obtain stable latent estimates of the internal thermal state, a Kalman filter is employed to fuse multimodal sensor inputs into a probabilistic state estimate \cite{sasiadek2002sensor}. The measurement-update step is expressed in Eq.~\ref{eq:kalman_update}:

\begin{equation}
\hat{x}_{t|t}
=
\hat{x}_{t|t-1}
+
K_t \bigl( y_t - H \hat{x}_{t|t-1} \bigr)
\label{eq:kalman_update}
\end{equation}

Here, $\hat{x}_{t|t}$ denotes the posterior state estimate at time $t$, while $\hat{x}_{t|t-1}$ is the prior predicted state obtained from the process model. The matrix $H$ specifies the observation model that maps latent thermal states to measurable quantities, and $K_t$ is the Kalman gain computed from the process and measurement noise covariances. The observation vector $y_t$ contains synchronized CBT, IMMU-based activity features, and UWB positional updates. Through recursive updates of $\hat{x}_{t|t}$, the Kalman filter produces a smoothed, noise-attenuated estimate of the latent thermal state, thereby improving the robustness of the DT model under sensor-level uncertainties. In any DT, robustness to external disturbances and parameter variations is essential; the Kalman filter addresses this requirement through its recursive state-estimation procedure, which continuously compensates for model–measurement discrepancies and attenuates the effect of process and sensor noise on the latent thermal state.

The state vector is defined as $x_t = [T_{\text{core}}, \dot{T}_{\text{core}}, A_{\text{level}}]^T$, representing core temperature, its rate of change, and activity level. The predicted prior $\hat{x}_{t|t-1}$ is generated by the nonlinear ODE and Markov models rather than a fixed linear state-transition matrix; the linear measurement update in Eq.~\ref{eq:kalman_update} then corrects this prior against the observed sensor data. The observation matrix $H$ is a selector matrix that maps the state components to their corresponding sensor channels in $y_t$. The process noise covariance $Q$ assumes continuous evolution of temperature with higher variance allowed for activity, while the measurement noise covariance $R$ is weighted by the inverse of each sensor's documented precision ($\sigma_{\text{CBT}} \approx 0.1^{\circ}\text{C}$). \x{The measurement-noise covariance $R$ is set from this logger precision, while the process-noise covariance $Q$ is initialized as a small diagonal matrix (of order $5\times10^{-3}$ on the temperature state, with a larger entry on the activity component) and adapted online from the recent residual variance. The Kalman gain $K_t$ is recomputed at each one-minute step from the current covariances.}

\subsubsection{Gaussian Process Learning}

To capture residual nonlinearities and characterize uncertainty in the digital-twin (DT) predictions, a Gaussian Process (GP) regression layer is applied to the residual errors between the measured and simulated CBT values \cite{seeger2004gaussian}. The GP prior is defined in Eq.~\ref{eq:gp_prior_definition}:

\begin{equation}
f(x) \sim \mathcal{GP}(m(x), k(x, x'))
\label{eq:gp_prior_definition}
\end{equation}

The covariance structure is modeled using a composite kernel, shown in Eq.~\ref{eq:gp_composite_kernel}, consisting of an RBF term that captures smooth physiological variations and a white-noise component representing high-frequency stochastic disturbances:

\begin{equation}
k = \sigma_{\text{const}}^2 k_{\text{RBF}} + k_{\text{white}}(\sigma_n^2)
\label{eq:gp_composite_kernel}
\end{equation}

In this formulation, $m(x)$ denotes the mean function and $k(x, x')$ the kernel function. Incorporating this GP layer enables explicit modeling of epistemic uncertainty due to incomplete sensor coverage and unmodeled biological dynamics, thereby providing calibrated confidence intervals for DT-based CBT predictions.

\x{For each animal, a dedicated GP is fit on its residual series $r(t) = T_{\text{cbt}}^{\text{meas}}(t) - T_{\text{cbt}}^{\text{DT}}(t)$, capturing individual-specific physiological deviations. Its inputs are the indoor THI and the IMMU-derived activity magnitude. The kernel hyperparameters (the constant scale $\sigma_{\text{const}}^2$, the RBF length-scale, and the white-noise variance $\sigma_n^2$ of Eq.~\ref{eq:gp_composite_kernel}) are obtained by type-II maximum likelihood, maximizing the log marginal likelihood of the residuals~\cite{rasmussen2006gaussian}.}





\subsubsection{DT Feature Generation}

The final stage of the digital-twin (DT) pipeline synthesizes a compact set of latent features that integrate information from both the physics-based thermal model and the probabilistic inference layer \cite{rao2025computational}. The resulting DT feature vector is defined in Eq.~\ref{eq:dt_feature_vector}:

\begin{equation}
F_{DT} = \left[ \hat{T}_{cbt},\; \hat{T}_{future},\; p_{stress},\; p_{behavior},\; \sigma_{uncertainty} \right]
\label{eq:dt_feature_vector}
\end{equation}

In this representation, $\hat{T}_{cbt}$ denotes the filtered CBT estimate from the DT model, while $\hat{T}_{future}$ corresponds to short-term thermal forecasts generated via sequential physiological simulation. The term $p_{stress}$ captures the probability of heat stress based on threshold-driven thermal risk inference, and $p_{behavior}$ encodes the inferred behavioral-state distribution. The quantity $\sigma_{uncertainty}$ represents posterior predictive uncertainty derived from the GP layer.

These DT-generated variables are incorporated into the multimodal learning pipeline, allowing the downstream ensemble model to reason jointly over observed sensor data and simulated physiological dynamics. This fusion of physical and data-driven representations forms the core of the proposed digital-twin–enhanced multimodal monitoring framework.

\subsection{Multi-Scale Feature Engineering}
\label{subsec:features}

Following multimodal alignment and DT simulation, the integrated dataset is transformed into a multi-scale feature representation that captures short-term physiological dynamics, mid-range behavioral patterns, and long-term environmental influences. This stage of processing aims to extract interpretable and temporally robust predictors that preserve both instantaneous fluctuations and aggregated contextual trends across modalities. The engineered features enable multimodal learning and improve generalization over time and across environments.

\subsubsection{Rolling Statistical Features}

To characterize localized variations and transient responses, rolling-window statistics are computed over each modality. Let $x_t$ denote a univariate sensor time series. The rolling statistical operator is defined in Eq.~\ref{eq:rolling_stats_definition}:
\begin{equation}
\bar{x} = \text{rolling}(x_t)\{\text{mean}, \text{std}, \text{max}, \text{min}, \text{var}, \text{skew}\}
\label{eq:rolling_stats_definition}
\end{equation}
Each statistic is computed within a fixed-length sliding window, producing multi-scale representations of recent temporal behavior. The mean and variance capture general magnitude and dispersion, while higher-order moments such as skewness and kurtosis characterize asymmetry and transient spikes. For instance, abrupt increases in $\text{std}(T_{cbt})$ may signal thermoregulatory instability, whereas elevated $\text{mean}(\text{activity})$ over short windows reflects restlessness or heat-induced movement.

\subsubsection{Temporal Encoding}

Animal behavior and physiological processes exhibit strong periodicity governed by circadian and management-related cycles. To encode these temporal regularities, each observation is augmented with cyclic time embeddings that preserve periodic continuity. The resulting temporal feature vector is defined in Eq.~\ref{eq:temporal_encoding_features}:
\begin{equation}
f_{\text{time}} = 
\begin{bmatrix}
\sin\left( \dfrac{2\pi h}{24} \right),\;
\cos\left( \dfrac{2\pi h}{24} \right),\;
\text{day\_of\_week}
\end{bmatrix}
\label{eq:temporal_encoding_features}
\end{equation}

These features allow the model to distinguish between similar physiological conditions occurring at different phases of the day, thereby capturing diurnal patterns in feeding, rumination, or thermoregulation. The sinusoidal encoding preserves continuity at day boundaries, avoiding discontinuities introduced by integer hour encoding.

\subsubsection{Physiological Derivatives}

Thermal regulation is inherently dynamic; hence, the rate of temperature change conveys critical information about metabolic adjustments \cite{farooq2010physiological}. To capture these short-term dynamics, first-order temporal derivatives of CBT are computed as shown in Eq.~\ref{eq:first_order_cbt_derivative}:
\begin{equation}
\frac{dT_{cbt}}{dt} = \text{diff}(T_{cbt})
\label{eq:first_order_cbt_derivative}
\end{equation}
Additionally, thermal differentials relative to the microclimatic environment are computed to quantify heat exchange gradients.This environmental temperature contrast is defined in Eq.~\ref{eq:thi_temperature_differential}:
\begin{equation}
\Delta d_{thi} = T_{cbt} - THI_{\text{indoor}}
\label{eq:thi_temperature_differential}
\end{equation}
Large positive differentials indicate physiological heat accumulation, while negative values suggest cooling phases, such as post-activity recovery or nighttime temperature decline.

\subsubsection{Cross-Modal Interaction Features}

Beyond single-modality patterns, physiological and behavioral states often emerge through coupled interactions among modalities. Cross-modal features are explicitly constructed to capture relational dependencies such as thermal-behavioral coupling and environmental compensation. The resulting interaction feature set is expressed in Eq.~\ref{eq:cross_modal_features}:
\begin{equation}
\begin{split}
F_{\text{cross}} = 
\{\, 
&THI_{\text{indoor}} - THI_{\text{outdoor}},\;
T_{\text{cbt}} \times \text{activity},\\
&\text{speed} \times \text{zone},\;
\ldots
\,\}
\end{split}
\label{eq:cross_modal_features}
\end{equation}

These interaction terms enhance the model’s sensitivity to joint conditions. For example, identifying heat stress when high $THI$ co-occurs with reduced movement or elevated CBT during resting states.

\subsubsection{Cumulative Heat Stress Metrics}

Chronic exposure to elevated internal temperature provides an early indication of physiological strain. To quantify long-term thermal load, a cumulative heat-stress metric is computed as shown in Eq.~\ref{eq:cumulative_heat_stress_metric}:
\begin{equation}
\tau_{\text{stress}} = \sum_t [T_{cbt}(t) > \tau]
\label{eq:cumulative_heat_stress_metric}
\end{equation}
where $\tau$ denotes the physiological heat-stress threshold. The cumulative duration $\tau_{\text{stress}}$ reflects the total time spent under thermal stress conditions, integrating transient episodes into a robust measure of sustained heat load.

\vspace{0.2em}
Together, these multi-scale statistical, temporal, physiological, and cross-modal features form a rich, interpretable representation of the cow’s thermophysiological state. They enable the subsequent learning components to discern patterns across timescales, ranging from minute-level fluctuations to daily stress accumulation, thereby improving the accuracy and stability of multimodal prediction.

\subsection{Base Models}
\label{subsec:basemodels}

The multimodal learning stage constitutes the first level of the hierarchical ensemble, in which each sensing modality is modeled independently to capture its intrinsic relationship with the target physiological variable. By training modality-specific learners, the system isolates domain-relevant predictive structures and mitigates feature redundancy across heterogeneous inputs. This modular architecture also enhances interpretability, allowing analysis of which physiological or environmental signals most strongly influence predictive performance. the collection of base learners is defined in Eq.~\ref{eq:base_models_def}:

\begin{equation}
M_{\text{base}} = { M_1, M_2, \ldots, M_n }
\label{eq:base_models_def}
\end{equation}
where each $M_i$ corresponds to an individual modality ( $M_{\text{cbt}}$, $M_{\text{uwb}}$, $M_{\text{immu}}$, $M_{\text{thi}}$, etc.), trained to map its feature subset $X_i$ to the physiological target $y$. All base models are implemented using the Light Gradient Boosting Machine (LightGBM), a decision-tree–based ensemble method optimized for high-dimensional and sparse time-series data. LightGBM constructs additive regression trees via gradient boosting, sequentially minimizing the empirical loss function , as expressed in Eq.~\ref{eq:lightgbm_loss}:

\begin{equation}
\theta_i = \arg\min_{\theta} \sum_{j=1}^{N_i} \mathcal{L}(y_j, f_\theta(X_{i,j}))
\label{eq:lightgbm_loss}
\end{equation}
where $\mathcal{L}$ denotes the regression loss (typically the squared error), $N_i$ is the number of samples for modality $i$, and $\theta$ represents the parameter set of the weak learners. Each model incrementally refines its predictions through stage-wise additive fitting, yielding modality-specific estimators $\hat{y}_i = M_i(X_i)$.

To assess and integrate the relative reliability of each modality, the coefficient of determination ($R^2$) from validation data is used as a quantitative performance indicator. Expert-based weights are computed to proportionally emphasize stronger modalities during the meta-model fusion stage. as shown in Eq.~\ref{eq:modality_weights}:

\begin{equation}
W_i = \frac{\max(0, R_i^2)}{\sum_{j=1}^{n} \max(0, R_j^2)}
\label{eq:modality_weights}
\end{equation}

These normalized weights reflect the explanatory contribution of each modality to the overall predictive task. Modalities with higher $R^2$ values, such as those capturing core physiological or thermoregulatory dynamics, receive greater influence in the subsequent fusion layer, while lower performing modalities are adaptively down-weighted.

This two fold process independent learning followed by performance-based weighting ensures that the ensemble leverages the complementary predictive strength of each modality without overfitting to any single data source. Consequently, the base modeling layer serves as a diverse ensemble of specialized “experts,” each capturing unique aspects of the cow’s behavioral, environmental, or physiological state.

\begin{figure*}[ht!]
\centering
\includegraphics[scale=0.235]{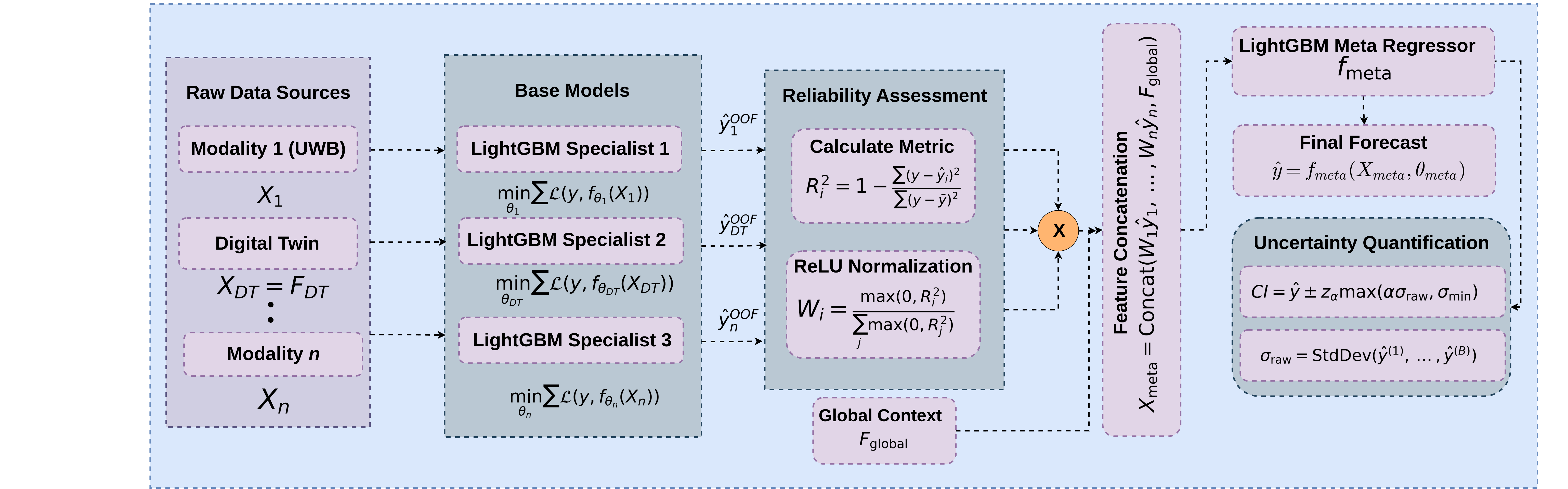}
\caption{The overview of the hierarchical fusion strategy. Modality-specific ``Specialist'' models ($M_1 \ldots M_n$) are trained independently. Their validation performance ($R^2$) determines their contribution weight ($W_i$). These weighted predictions are concatenated with global context features ($F_{global}$) to train the Meta-Regressor ($f_{meta}$), which outputs both the final forecast and uncertainty bounds.}
\label{fig:Stacked Ensem}
\end{figure*}

\subsection{Meta-Model}

The second stage of the proposed architecture performs ensemble-level fusion of modality-specific predictions through a meta-learning framework. As illustrated in Figure \ref{fig:Stacked Ensem}, the proposed framework employs a hierarchical fusion strategy to synthesize heterogeneous sensor inputs. The process begins by training isolated LightGBM `specialists' on distinct feature sets. For instance, UWB, DT features, and environmental data, to capture domain-specific signals without interference. To ensure robustness against sensor failure, the system applies an adaptive reliability assessment where the Out-Of-Fold (OOF) predictions of each specialist are evaluated; validation $R^{2}$ scores are processed through a ReLU-based normalization to derive dynamic weights ($W_{i}$) that effectively suppress unreliable models. These weighted predictions are subsequently concatenated with global temporal context ($F_{global}$) to form the comprehensive feature vector $X_{meta}$. The LightGBM Meta-Regressor ($f_{meta}$) processes this fused representation to generate the Core Body Temperature forecast, while a parallel uncertainty quantification module utilizes the variance of bootstrapped replicas ($\sigma_{raw}$) to construct calibrated confidence intervals ($CI$).

To prevent data leakage during the training of the meta-learner, we
employed a nested cross-validation scheme. The base model predictions $\hat{y}_i$
used in Eq.~\ref{eq:meta_input} were generated using Out-of-Fold (OOF) predictions.
Specifically, for every fold $k$ in the training set, base models were trained on
the remaining $k - 1$ folds and predicted on fold $k$. This ensures the meta-
model learns to correct the base models' unseen-data errors, rather than
overfitting to their training residuals.

The input to the meta-model is constructed by concatenating the expert-weighted predictions from all base learners along with auxiliary global features, as defined in Eq.~\ref{eq:meta_input}:

\begin{equation}
X_{\text{meta}} = \text{Concat}(w_1\hat{y}_1, w_2\hat{y}_2, \dots, w_n\hat{y}_n, F_{\text{global}})
\label{eq:meta_input}
\end{equation}

%

where $\hat{y}_i$ denotes the predicted output from the $i$-th modality-specific model, $w_i$ represents its expert-derived weight based on performance reliability, and $F_{\text{global}}$ includes contextual features such as temporal encodings, environmental indicators, and productivity statistics. This concatenated representation captures both intra-modal and cross-modal dependencies, enabling the meta-learner to integrate heterogeneous physiological, behavioral, and environmental information effectively.

The meta-model is implemented using a LightGBM regressor \cite{ke2017lightgbm} parameterized by $\theta_{\text{meta}}$, which applies gradient boosting over decision trees to model nonlinear relationships between the fused feature representation and the target output. The prediction function is expressed in Eq.~\ref{eq:meta_prediction}:

\begin{equation}
\hat{y} = f_{\text{meta}}(X_{\text{meta}}, \theta_{\text{meta}})
\label{eq:meta_prediction}
\end{equation}
This optimization is guided by validation-based hyperparameter tuning using the Optuna framework \cite{akiba2019optuna}, ensuring robust generalization and balanced feature utilization across modalities.

By learning hierarchical relationships among the modality-level predictions, the meta-model effectively captures synergistic dependencies, such as how environmental heat load and physical activity jointly influence thermoregulation. Furthermore, this two-tier ensemble structure mitigates overfitting to any individual sensor stream by allowing the meta-model to weigh and recalibrate the contribution of each modality according to its reliability and contextual relevance. Consequently, the final ensemble output $\hat{y}$ represents a globally optimized, uncertainty-aware \x{forecast} of the physiological variable of interest, derived from the collective intelligence of multimodal base learners.

\subsection{Uncertainty Quantification}

To promote interpretability, reliability, and decision-level transparency, the proposed framework incorporates an explicit uncertainty estimation mechanism that quantifies the confidence of each prediction. This component addresses the intrinsic variability of multimodal data and the epistemic uncertainty arising from limited training samples or imperfect sensor coverage. The method estimates uncertainty by resampling ensembles that approximate the predictive distribution through repeated random perturbations of the training data.

\subsubsection{Bootstrap Sampling}

Epistemic uncertainty, which reflects the model’s lack of knowledge about unseen regions of the input space, is captured using bootstrap resampling \cite{dixon2006bootstrap}. Multiple independent replicas of the meta-model are trained on randomly resampled subsets of the training data, as defined in Eq.~\ref{eq:bootstrap_sampling}:
\begin{equation}
\left\{\hat{y}^{(b)}\right\}_{b=1}^{B} = \text{Bootstrap}(X, y)
\label{eq:bootstrap_sampling}
\end{equation}
where $B$ denotes the number of bootstrap iterations, and $\hat{y}^{(b)}$ represents the prediction generated by the $b$-th replica. Each replica is trained on a slightly perturbed version of the dataset, effectively exploring different regions of the hypothesis space. This process yields a distribution of predictions at each time step, approximating the epistemic uncertainty inherent in the model’s decision boundaries.

\subsubsection{Variance-Based Uncertainty Estimation}

The dispersion among the bootstrapped predictions is used to quantify predictive variance. The raw standard deviation of the ensemble outputs, denoted $\sigma_{\text{raw}}$, is scaled by a calibration coefficient $\alpha$ and bounded by a minimum uncertainty threshold $\sigma_{\min}$ to prevent degenerate confidence intervals, as defined in Eq.~\ref{eq:variance_uncertainty}:
\begin{equation}
\sigma_{\text{final}} = \max(\alpha \cdot \sigma_{\text{raw}}, \sigma_{\min})
\label{eq:variance_uncertainty}
\end{equation}
This scaling procedure normalizes the uncertainty estimates across samples and prevents underestimation in highly confident but data-sparse regions. The factor $\alpha$ is empirically determined through coverage-based calibration using a validation set, aligning the nominal and empirical confidence levels of the prediction intervals.

\subsubsection{Prediction Interval Construction}

For each prediction $\hat{y}$, the calibrated uncertainty $\sigma_{\text{final}}$ is used to derive a symmetric confidence interval corresponding to the desired significance level, as defined in Eq.~\ref{eq:prediction_interval_construction}:

\begin{equation}
CI = \hat{y} \pm z_\alpha \sigma_{\text{final}}
\label{eq:prediction_interval_construction}
\end{equation}
where $z_\alpha$ is the standard normal quantile corresponding to the chosen confidence level. The resulting interval provides an interpretable probabilistic bound around the model’s point estimate, indicating the range within which the true physiological value is expected to lie with a specified degree of confidence.

\vspace{0.2em}
The incorporation of bootstrap-based uncertainty quantification serves two critical functions: (1) it provides a principled measure of predictive confidence for decision support in heat-stress detection, and (2) it enables model evaluation via coverage metrics such as Prediction Interval Coverage Probability (PICP). By quantifying both prediction accuracy and reliability, the framework enhances the transparency and operational trustworthiness of multimodal digital-twin monitoring systems.

\subsection{Prediction Heads}

The proposed multimodal digital-twin framework culminates in a dual-output prediction stage that simultaneously performs continuous regression of core body temperature (CBT) and binary classification of heat-stress states. These complementary prediction heads are designed to jointly provide quantitative physiological forecasts and actionable stress indicators, enabling both analytical interpretation and real-time decision support for herd management systems.



\subsubsection{Continuous CBT Prediction}

The first prediction head is designed to \x{forecast} the future trajectory of a cow's core body temperature (CBT) with high temporal precision. The regression head is explicitly designed to forecast the point-wise Core Body
Temperature (CBT) at the forecast horizon \( h = 120 \) minutes
(\( \hat{y}_{t+h} \)). Unlike aggregate metrics (such as window averages or
maximums), which mask temporal dynamics, point-wise prediction forces the
digital twin to maintain phase-accurate synchronization with the animal's
thermoregulatory cycle. Simultaneously, the classification head (Section \ref{section : heat}) predicts the probability that the {maximum CBT within the $[t, t + 120]$ window} will exceed the heat stress threshold ($\theta$). This dual-target approach allows for precise point forecasting while capturing risk events that might occur between sample points.

Given observations up to time $t$, the regression model predicts the specific temperature at forecast horizon $h = 120$ min, as defined in Eq.~\ref{eq:cbt_forecast}:

\begin{equation}
\hat{y}_{t+h} = f(X_t, X_{t-1}, \dots, X_{t-p}) 
\label{eq:cbt_forecast}
\end{equation}

Here, $X_t$ denotes the multimodal feature vector at time $t$, and $p$ represents the temporal lag that captures the autoregressive context window. \x{The autoregressive context is constructed from multi-scale rolling windows of $\{15,\,60,\,240\}$ minutes, giving an effective lookback $p$ that spans sub-hourly fluctuations through a four-hour history. This range covers the dominant timescales of bovine thermoregulation, including the environmental-to-CBT thermal lag of roughly $1.5$--$3$~hours~\cite{islam2023revealing} captured by the four-hour window.}

\x{The two-hour forecast horizon ($h = 120$~min) is motivated by the physiology of heat accumulation. Ambient heat load drives core body temperature with an intrinsic delay of roughly $1.5$--$3$~hours~\cite{islam2023revealing}; a two-hour horizon therefore falls within this lag window and corresponds to the interval over which a cooling intervention (forced ventilation, sprinklers, or shade) can take effect before CBT crosses the threshold. Shorter horizons provide insufficient lead time for intervention, while longer horizons extend beyond the predictive signal available from current conditions~\cite{17}.} The resulting forecast $\hat{y}_{t+h}$ is paired with an uncertainty-aware confidence interval derived from the bootstrap calibration process, as shown in Eq.~\ref{eq:cbt_ci}:

\begin{equation}
\hat{y}_{\text{future}} \pm CI 
\label{eq:cbt_ci}
\end{equation}

This formulation provides both a point prediction and a probabilistic confidence bound, enabling robust thermal trend forecasting and early detection of deviations indicative of heat stress or abnormal physiological responses.

\subsubsection{Heat-Stress Classification}
\label{section : heat}

The second prediction head performs binary classification to determine whether a cow is under heat-stress conditions. The classification probability is computed via a logistic (sigmoid) activation applied to the predicted CBT value, as shown in Eq.~\ref{eq:stress_probability}:
\begin{equation}
P(\text{stress}) = \frac{1}{1 + e^{-\beta(\hat{y} - \theta)}}
\label{eq:stress_probability}
\end{equation}

where $\theta$ is the clinical heat stress threshold (38.8°C), $\hat{y}$ is the predicted CBT, and $\beta$ {is a calibrated scaling parameter} that controls the steepness of the decision boundary. A higher $\beta$ simulates a sharper transition into heat stress, while a lower $\beta$ accounts for boundary uncertainty. The corresponding binary class label is obtained by thresholding the predicted CBT value, as defined in Eq.~\ref{eq:stress_decision}:
\begin{equation}
\hat{c} = [\hat{y} > \theta]
\label{eq:stress_decision}
\end{equation}
Here, $\hat{c}=1$ indicates a heat-stressed condition, whereas $\hat{c}=0$ denotes a thermally normal state.

\vspace{0.3em}
This dual-headed design unifies quantitative forecasting and categorical decision-making. The regression head models fine-grained thermal variations, while the classification head provides actionable stress detection for precision livestock monitoring. Together, they enable proactive physiological management and interpretable, real-time assessment of animal well-being. The complete computational pipeline, integrating both the physics-informed DT simulation and the stacked ensemble learning strategy, is described in Algorithm ~\ref{alg:mmcows_pipeline}.

\begin{algorithm}[t]
\footnotesize
\setlength{\tabcolsep}{1pt}
\caption{Proposed Training Algorithm for CBT And Stress Prediction}
\label{alg:mmcows_pipeline}

\begin{minipage}{\columnwidth}
\begin{algorithmic}[1]

\Require Sensor data $D_s$, weather $D_w$, milk $D_m$
\Ensure $\hat{T}_{cbt}$, $CI$, $P_{stress}$

\State \textbf{1. Data Acquisition \& Prep}
\State $D_{raw} \gets \{D_{uwb}, D_{imu}, D_{ankle}, D_{cbt}, D_{thi}, D_w, D_m\}$

\State $D_{proc} \gets 
\textsc{Merge}\big(
  \textsc{Group}(
    \textsc{Clean}(
      \textsc{Align}(D_{raw})
    )
  ),\;
  D_w
\big)$

\State \textbf{2. Digital Twin Simulation}

\State $A \gets \textsc{ThermalODE}()$
\State $B \gets \textsc{MarkovModel}()$
\State $M_{sys} \gets \textsc{KalmanFusion}(A, B)$

\State $M_{gp} \gets \textsc{GPLearn}(D_{cbt}^{hist}, D_{thi}^{hist})$

\State $D_{DT} \gets 
\textsc{FeedbackLoop}\big(
  \textsc{GenFeatures}(M_{sys}, M_{gp}, D_{proc})
\big)$

\State \textbf{3. Feature Engineering}
\State $F \gets \{\textsc{RollStats}, \textsc{TempEnc}, \textsc{CrossMod}, \textsc{CumStress}\}$
\State $X \gets \textsc{Concat}(f(D_{DT})\ \forall f \in F)$

\State \textbf{4. Stacked Ensemble Learning}
\State $(X_{tr}, X_{val}) \gets \textsc{GroupKFold}(X)$

\For{modality $m$}
  \State $M_m \gets \textsc{LGBM}(X_{m,tr}, y_{tr})$
  \State $\hat{y}_m \gets M_m(X_{m,val})$
\EndFor

\State $W \gets \textsc{CalcWeights}(R^2_m)$
\State $X_{meta} \gets [\hat{y}_1{:}\hat{y}_m, W, X_{glob}]$
\State $M_{meta} \gets \textsc{LGBM}(X_{meta}^{val}, y_{val})$

\State \textbf{5. Inference \& Uncertainty}
\State $\hat{T}_{cbt} \gets M_{meta}(X_{meta}^{test})$
\State $\sigma \gets \textsc{Bootstrap}(M_{meta})$
\State $CI \gets \hat{T}_{cbt} \pm 1.96\sigma$
\State $P_{stress} \gets \sigma(\alpha(\hat{T}_{cbt} - \theta_{stress}))$

\Return $\hat{T}_{cbt},\; CI,\; P_{stress}$

\end{algorithmic}
\end{minipage}
\end{algorithm}

\section{Experimental Details}
\label{sec:experimental_setup}

This section presents the experimental setup used to evaluate the proposed method to provide the basis for the results and analysis presented in the following sections. We first introduce the dataset and its key properties. Then, we outline the evaluation metrics used to measure both regression performance and clinical classification robustness. 

\subsection{Dataset}
\label{subsec:datasets}

This subsection introduces the {MmCows} benchmark dataset, a multimodal dataset specifically designed for precision livestock farming \cite{vu2024mmcows}. This dataset is important due to its synchronized physiological and behavioral sensing data, but it also poses significant challenges such as data scarcity, sensor noise, and complex temporal dynamics. The dataset is described in terms of its structure, sensor modalities, and statistical distribution, highlighting how the proposed {digital twin} framework is tailored to achieve high performance under these low-data regimes. The MmCows dataset is a comprehensive multimodal dataset collected from intensively monitored Holstein dairy cows over a continuous period at the University of Wisconsin-Madison. Unlike traditional datasets that rely solely on visual observation, MmCows integrates wearable sensing, environmental logs, and physiological ground truth.

The data collection involved continuous monitoring using neck-mounted tags (providing Ultra-Wideband location and IMU acceleration), ankle pedometers, and vaginal temperature loggers. The dataset was resampled to a 1-minute resolution for this study. 

In the dataset, while behavioral data (acceleration) is abundant, high-stress events (CBT $> 38.8^\circ$C) are relatively rare, constituting a ``tail'' in the distribution that requires robust modeling. Additionally, the dataset poses cross-modality inconsistencies, where environmental heat (THI) does not instantly translate to body heat due to thermal inertia. These characteristics make heat stress prediction on MmCows particularly difficult. To address these issues, our proposed hybrid framework incorporates a Physics-Informed DT and an Expert-based Fusion mechanism to robustly integrate these heterogeneous cues.

To provide a comprehensive and fair assessment of model performance on heat stress prediction, we selected widely used evaluation metrics, focusing on both regression accuracy and clinical decision capability. These metrics include Mean Absolute Error (MAE), Root Mean Squared Error (RMSE), Coefficient of Determination ($R^2$), Prediction Interval Coverage Probability (PICP), and clinical classification scores. Together, they reflect both the precision of the temperature forecast and the reliability of the system for uncertainty quantification.

\subsection{Cross-Validation Protocol}
\x{Model performance is assessed with a grouped, nested cross-validation scheme that uses the animal identifier (\texttt{cow\_id}) as the grouping variable, ensuring that no cow appears in both the training and the test partition. The outer GroupKFold uses $K = 10$ folds. Within each fold, an outer split defines a held-out \emph{test} fold that is withheld until the final evaluation; out-of-fold (OOF) predictions of the modality-specific base learners are generated on the outer-training cows, on which the expert $R^2$ weights $W_i$ are computed; the LightGBM meta-model and its Optuna hyperparameter search are fit and selected on the inner validation split; and the uncertainty calibration ($\alpha$ and $\sigma_{\min}$ in Eq.~\ref{eq:variance_uncertainty}) is estimated on inner validation and evaluated on the outer test fold. No test-fold cow therefore contributes to feature scaling, digital-twin parameter fitting, expert-weight computation, hyperparameter optimization, or uncertainty calibration. This grouped design is consistent with recommended practice for data with hierarchical (per-animal) and temporal structure~\cite{roberts2017cross,bergmeir2012use}.}

\subsection{Evaluation Metrics}

To provide a comprehensive assessment of the proposed framework, performance is evaluated across three dimensions: the accuracy of continuous Core Body Temperature (CBT) forecasting, the reliability of uncertainty quantification, and the robustness of clinical heat-stress detection. For the regression task, we utilize Mean Absolute Error (MAE), Root Mean Squared Error (RMSE), and the Coefficient of Determination ($R^2$) to quantify predictive precision. Specifically, MAE provides an interpretable average error in $^\circ\text{C}$, RMSE penalizes large deviations that are critical near the onset of heat stress, and $R^2$ captures the explained variance across animals with heterogeneous thermoregulatory responses. To validate the model's confidence estimates, we employ the Prediction Interval Coverage Probability (PICP), ensuring that the generated uncertainty bounds are statistically calibrated. Clinical decision robustness is assessed using threshold-based classification metrics. Because heat-stress events are comparatively less frequent than normal states, we emphasize the F1-score (precision–recall balance) and AUC (threshold-independent discrimination), and report accuracy only as a secondary measure.
\begin{table*}[ht!]
\centering
\captionsetup{justification=centering}

\caption{Performance Comparison of Baseline Models and the Proposed Multimodal Model.}
\label{tab:baseline_comparison}
\scriptsize
\begin{tabular}{lccc}
\toprule
{Model} & {MAE} $\downarrow$ & {RMSE} $\downarrow$ & {$R^2$} $\uparrow$ \\
\midrule
Linear Regression & 0.1737 $\pm$ 0.0293 & 0.2112 $\pm$ 0.0323 & 0.4571 $\pm$ 0.1688 \\
Random Forest & 0.1712 $\pm$ 0.0125 & 0.2206 $\pm$ 0.0181 & 0.4184 $\pm$ 0.0669 \\
Decision Tree & 0.1576 $\pm$ 0.0171 & 0.2021 $\pm$ 0.0262 & 0.5025 $\pm$ 0.1393 \\
Gradient Boosting & 0.1552 $\pm$ 0.0122 & 0.1987 $\pm$ 0.0210 & 0.5242 $\pm$ 0.0984 \\
XGBoost & 0.1538 $\pm$ 0.0120 & 0.1972 $\pm$ 0.0203 & 0.5909 $\pm$ 0.0954 \\
SVR & 0.1793 $\pm$ 0.0227 & 0.2259 $\pm$ 0.0290 & 0.3908 $\pm$ 0.1130 \\
KNN & 0.2139 $\pm$ 0.0270 & 0.2688 $\pm$ 0.0350 & 0.1438 $\pm$ 0.1127 \\
LSTM & 0.1826 $\pm$ 0.0171 & 0.2309 $\pm$ 0.0204 & 0.3672 $\pm$ 0.0473 \\
Temporal Fusion Transformer & 0.1892 $\pm$ 0.0224 & 0.2391 $\pm$ 0.0271 & 0.3165 $\pm$ 0.1215 \\
\midrule
Proposed Model & 0.114 & 0.143 & 0.7835 \\
\bottomrule
\end{tabular}
\end{table*}

\paragraph{Mean Absolute Error (MAE)}
The Mean Absolute Error (MAE) quantifies the average magnitude of errors in a set of predictions, without considering their direction, as defined in Eq. ~\ref{eq:mae} :

\begin{equation}
    MAE = \frac{1}{N} \sum_{i=1}^{N} |y_i - \hat{y}_i| 
    \label{eq:mae}
\end{equation}

where $N$ is the total number of samples, $y_i$ is the ground truth Core Body Temperature (CBT), and $\hat{y}_i$ is the predicted CBT.

\paragraph{Root Mean Square Error (RMSE)}
RMSE is a quadratic scoring rule that also measures the average magnitude of the error. It gives a relatively high weight to large errors, making it useful for identifying when the model completely misses a heat stress spike, as defined in Eq. ~\ref{eq:rmse} :

\begin{equation}
    RMSE = \sqrt{\frac{1}{N} \sum_{i=1}^{N} (y_i - \hat{y}_i)^2}
    \label{eq:rmse}
\end{equation}

\paragraph{Coefficient of Determination ($R^2$)}

The $R^2$ score provides an indication of the goodness of fit and therefore a measure of how well unseen samples are likely to be predicted by the model, as defined in Equation~\ref{eq:r2} :

\begin{equation}
    R^2 = 1 - \frac{\sum_{i=1}^{N} (y_i - \hat{y}_i)^2}{\sum_{i=1}^{N} (y_i - \bar{y})^2}
    \label{eq:r2}
\end{equation}

where $\bar{y}$ is the mean of the observed data. An $R^2$ of 1.0 indicates perfect prediction.

\paragraph{Prediction Interval Coverage Probability (PICP)}
To evaluate the uncertainty quantification of our model, specifically the 95\% confidence intervals generated by the ensemble, we use PICP. It measures the percentage of ground truth values that fall within the predicted upper ($\hat{y}_{upper}$) and lower ($\hat{y}_{lower}$) bounds, as defined in Eq. ~\ref{eq:picp} :

\begin{equation}
    PICP = \frac{1}{N} \sum_{i=1}^{N} \mathbb{I}(\hat{y}_{lower, i} \le y_i \le \hat{y}_{upper, i})
    \label{eq:picp}
\end{equation}

where $\mathbb{I}$ is the indicator function. A robust model should achieve a PICP close to the nominal confidence level (95\%).

\paragraph{Clinical Classification Metrics}
While the model is trained as a regressor, its practical utility lies in detecting heat stress. We threshold the predicted CBT at {38.8$^\circ$C} (the clinical threshold for heat stress) to calculate AUC and F1-Score, as defined in Eq. ~\ref{eq:f1} : \x{This $38.8^{\circ}$C threshold defines the ground-truth heat-stress events for the classification metrics; it corresponds to the thermoneutral vaginal temperature of dairy cattle ($38.8^{\circ}$C; rectal $38.5^{\circ}$C)~\cite{oliveira2025heat} and is measured by the MmCows intravaginal logger. It is distinct from the classifier operating point (Section~\ref{sec:results}, Fig.~\ref{fig:classification}), which is selected on the predicted-CBT axis.}

\begin{equation}
    F1 = 2 \cdot \frac{Precision \cdot Recall}{Precision + Recall}
    \label{eq:f1}
\end{equation}

These metrics ensure the model is evaluated not just on numerical fit, but on its ability to alert farmers to at-risk animals.

\section{Results and Analysis}
\label{sec:results}

This section presents a comprehensive evaluation of our proposed multimodal framework for dairy cattle monitoring, leveraging the MmCows dataset. We first benchmark our complete model against a suite of established ML baselines to establish its superiority. Subsequently, we conduct a series of ablation studies to dissect the contribution of individual data modalities and to quantify the specific impact of our novel physics-informed DT component. A qualitative analysis is provided regarding the model's predictive performance, error characteristics, and reliability for heat-stress classification.

\begin{figure}[t!]
    \centering
    \includegraphics[scale=.12]{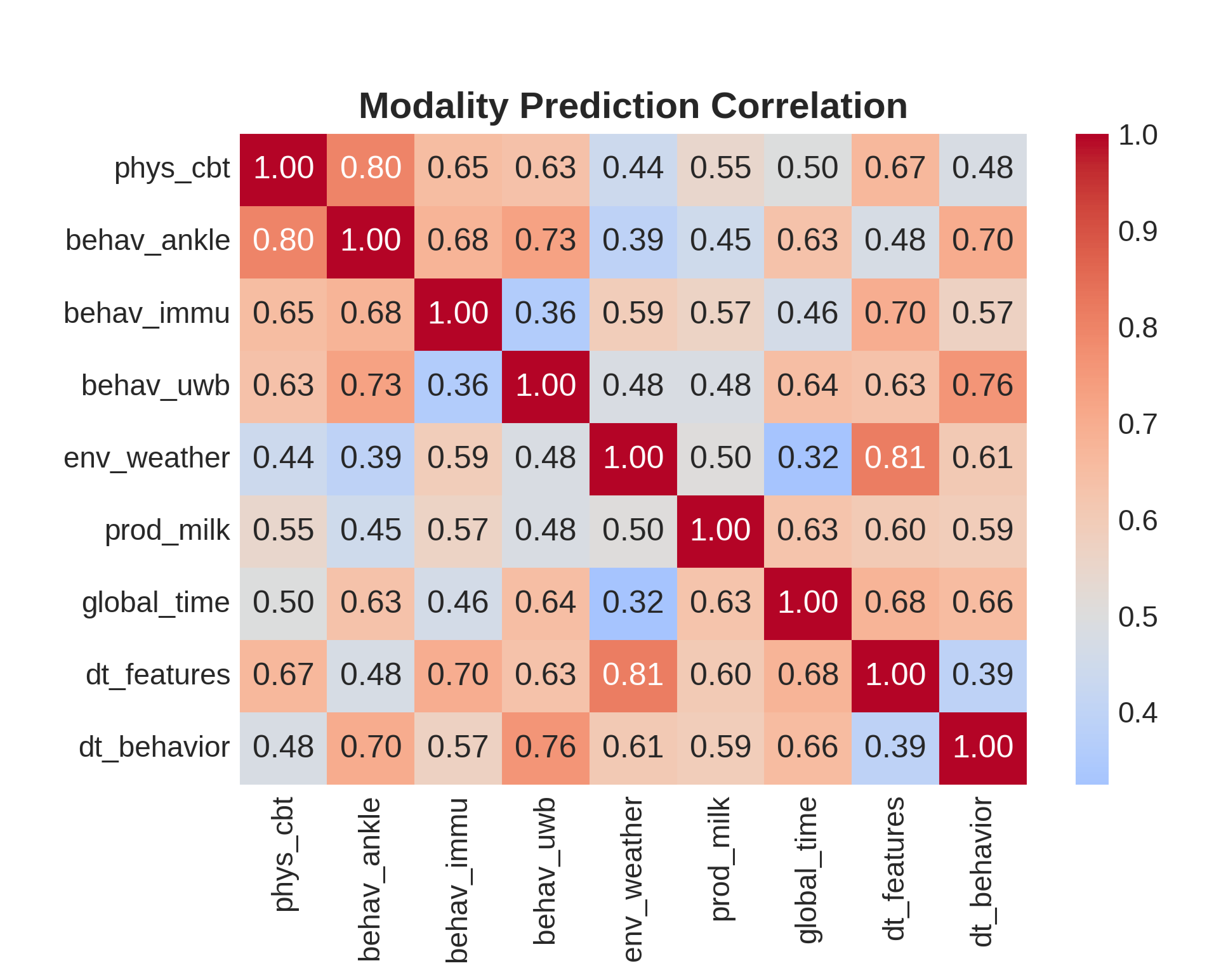} 
    
    \caption{{Modality Prediction Correlation Matrix.} This heatmap illustrates the Pearson correlation coefficients between the predictions of the modality-specific base learners. The high correlation between \texttt{dt\_features} and \texttt{phys\_cbt} (0.67) validates the DT's physiological fidelity, while lower correlations in environmental features confirm the model diversity required for the stacked ensemble.}
    
    \label{fig:correlation_matrix}
\end{figure}

\subsection{Benchmarking Against Baseline Models}
The performance of our proposed multimodal model was compared against nine baseline models, encompassing linear models, tree-based ensembles, and deep learning architectures designed for time-series data. To ensure fair benchmarking, input structures were tailored to each architecture. While all models utilized the standardized feature set (Section \ref{subsec:features}), deep sequence models (LSTM, TFT) were trained on multivariate time-series tensors (Batch $\times$ Window $\times$ Features) rather than flattened vectors. We utilized these engineered signals instead of raw sensor streams to mitigate convergence issues caused by high sensor noise, ensuring the deep baselines were evaluated at their optimal capability.

As presented in Table~\ref{tab:baseline_comparison}, XGBoost emerged as the best-performing baseline model, achieving an $R^2$ of 0.5909. This is consistent with its strong performance on tabular data with complex feature interactions. However, our proposed multimodal model dramatically outperformed all baselines across every metric. It achieved an $R^2$ of {0.783}, representing a remarkable improvement in explained variance over the best baseline. The MAE and RMSE were also significantly reduced to 0.114 and 0.143, respectively, indicating not only better average accuracy but also fewer large, costly prediction errors.

The derived classification metrics further highlight the model's practical utility. For the critical task of heat stress prediction, the model achieved an AUC of 0.967, an accuracy of 89.38\%, and an F1-score of 0.8425. This demonstrates that the model is not only accurate in regression but also highly effective for binary decision-making tasks, such as triggering automated cooling systems or generating veterinarian alerts.

\begin{figure*}[ht!]
\centering
\includegraphics[scale=0.088]{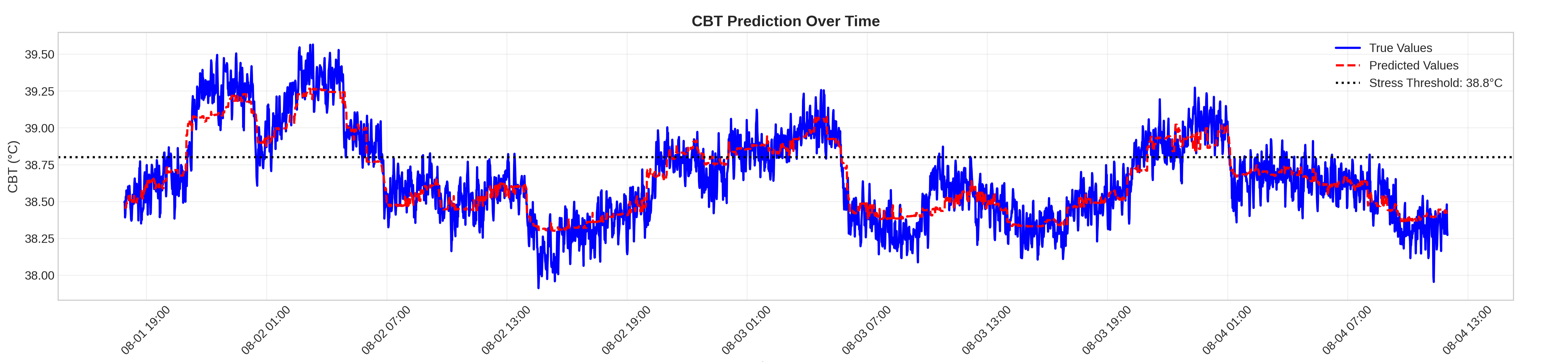}
    \caption{CBT Prediction Over Time. The model's predictions (red dashed line) accurately track the true CBT values (blue solid line) and effectively capture the critical crossing of the heat stress threshold (black dotted line at 38.8$^\circ$C). This demonstrates the model's capability for timely early warnings.}
    \label{fig:cbt_time_series}
\end{figure*}

\begin{table*}[htbp]
\centering
\captionsetup{justification=centering}

\caption{Ablation Study on Feature Group Importance.}
\label{tab:ablation_modality}
\scriptsize
\begin{tabular}{lccc}
\toprule
{Feature Group} & {MAE} $\downarrow$ & {RMSE} $\downarrow$ & {$R^2$} $\uparrow$ \\
\midrule
phys\_cbt & 0.1830 $\pm$ 0.0145 & 0.2313 $\pm$ 0.0190 & 0.3786 $\pm$ 0.0644 \\
behav\_ankle & 0.2340 $\pm$ 0.0159 & 0.2925 $\pm$ 0.0155 & 0.3459 $\pm$ 0.0330 \\
behav\_immu & 0.2349 $\pm$ 0.0177 & 0.2923 $\pm$ 0.0176 & 0.2578 $\pm$ 0.0422 \\
behav\_uwb & 0.2351 $\pm$ 0.0195 & 0.2943 $\pm$ 0.0217 & 0.1429 $\pm$ 0.0179 \\
env\_weather & 0.2133 $\pm$ 0.0112 & 0.2600 
$\pm$ 0.0136 & 0.2147 $\pm$ 0.0321 \\
prod\_milk & 0.2372 $\pm$ 0.0144 & 0.2978 $\pm$ 0.0159 & 0.0513 $\pm$ 0.0369 \\
global\_time & 0.2014 $\pm$ 0.0126 & 0.2529 $\pm$ 0.0145 & 0.2563 $\pm$ 0.0444 \\
dt\_features & 0.2342 $\pm$ 0.0155 & 0.2929 $\pm$ 0.0148 & 0.3520 $\pm$ 0.0397 \\
\midrule
Proposed Model (All) & 0.114 $\pm$ 0.0109 & 0.143 $\pm$ 0.0140 & 0.7835 $\pm$ 0.0499 \\
\bottomrule
\end{tabular}
\end{table*}

\subsection{Ablation Study on Modality Contribution}

To understand the synergistic value of the multimodal approach, we performed an ablation study where the proposed model was trained using only one feature group at a time. This isolates the predictive power of each data modality.

The results in Table~\ref{tab:ablation_modality} are profoundly revealing. No single modality was sufficient for accurate prediction. The best individual feature group, `phys\_cbt` (current and historical CBT), achieved a modest $R^2$ of 0.3786. Most other modalities performed poorly, with $R^2$ values near zero. This indicates that models trained solely on this data performed worse than a naive model that simply predicts the mean CBT for all inputs. As illustrated in (Figure~\ref{fig:correlation_matrix}), the Pearson correlation coefficients between the predictions of the modality-specific base learners reveal varying degrees of dependency, confirming the model diversity required for the stacked ensemble. This underscores a central insight of our work: {the predictive power is not inherent in any single sensor stream but emerges from their intelligent fusion}. The model successfully learns complex, non-linear interactions between physiology, behavior, and the environment that are imperceptible when analyzing modalities in isolation.

\subsection{Impact of the Physics-Informed Digital Twin}
To isolate the contribution of DT component, we compared the performance of the full proposed model with an identical version where all DT-derived features ( `dt\_cbt\_prediction`, `dt\_stress\_probability`) were removed.

\begin{table}[ht!]
\centering
\scriptsize
\caption{Ablation Study on the Digital Twin Component.}
\label{tab:ablation_dt}
\begin{tabular}{lccc}
\toprule
{Metric} & {With DT} & {Without DT} & {Improvement} \\
\midrule
MAE & 0.1147 & 0.1385 & +17.22\% \\
RMSE & 0.1435 & 0.1756 & +18.29\% \\
$R^2$ & 0.7835 & 0.6927 & +13.06\% \\
PICP & 92.38\% & 86.52\% & +6.77\% \\
\bottomrule
\end{tabular}
\end{table}

The inclusion of the DT consistently enhanced performance across all key metrics, as shown in Table~\ref{tab:ablation_dt}. The $R^2$ value dropped by 13.06\% without the DT, and the prediction errors (MAE/RMSE) increased by over 17\%. This confirms that the DT provides a significant boost to predictive accuracy. More importantly, the {Prediction Interval Coverage Probability (PICP)}, a measure of the model's uncertainty quantification reliability, fell from 92.38\% to \x{86.52\%}. This demonstrates that the DT not only improves point-estimate accuracy but also provides a more reliable and trustworthy measure of prediction confidence. By grounding the data-driven model in the physics of thermoregulation, the DT acts as a powerful regularizer, constraining predictions to be physically plausible and improving generalization.

\begin{figure*}[ht!]
    \centering
    \includegraphics[scale=.4]{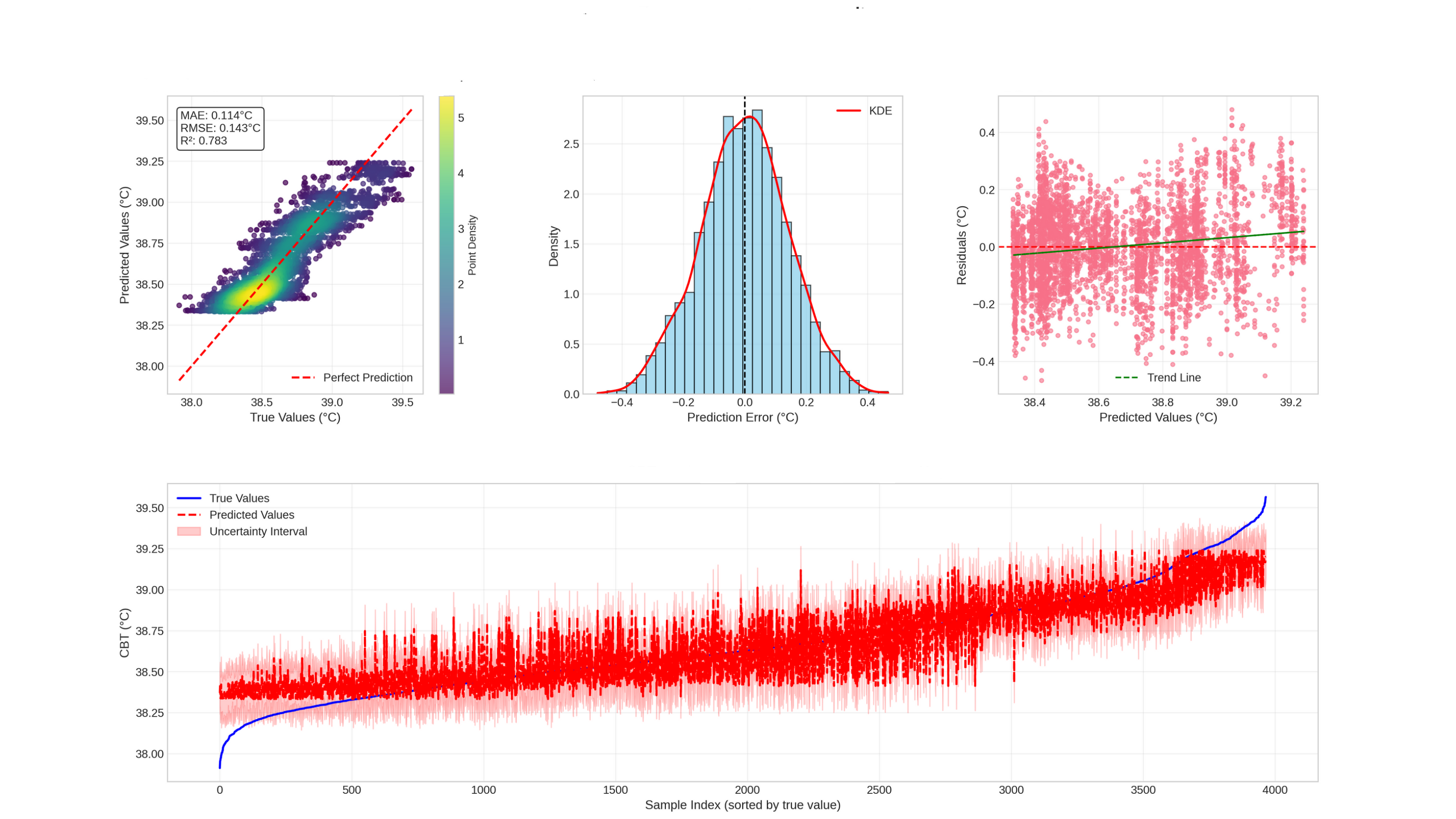}
    \caption{Comprehensive Regression Analysis. (Top-left) Prediction vs True Values scatter plot with color gradient indicating prediction density and red dashed line representing perfect prediction. (\x{Top-middle}) Error Distribution histogram with KDE overlay showing the distribution of prediction errors. (\x{Top-right}) Residual Distribution plot with trend line indicating no systematic bias. (\x{Bottom}) True Values and Predictions over time with uncertainty intervals (shaded area), demonstrating the model's ability to quantify prediction uncertainty. The model achieves MAE: 0.114°C, RMSE: 0.143°C, and R²: 0.783.}
    \label{fig:regression_analysis}
\end{figure*}

\subsection{Regression Performance Analysis}

Figure~\ref{fig:cbt_time_series} visualizes the model's predictive performance over a 3-day period. The predicted CBT values (red dashed line) closely track the true values (blue solid line), even during periods of rapid fluctuation. Critically, the model successfully anticipates and follows the trajectory as the cow's CBT crosses the heat stress threshold (black dotted line at 38.8$^\circ$C), demonstrating its capability for timely early warnings.

\begin{figure*}[ht!]
    \centering
    \includegraphics[scale=0.15]{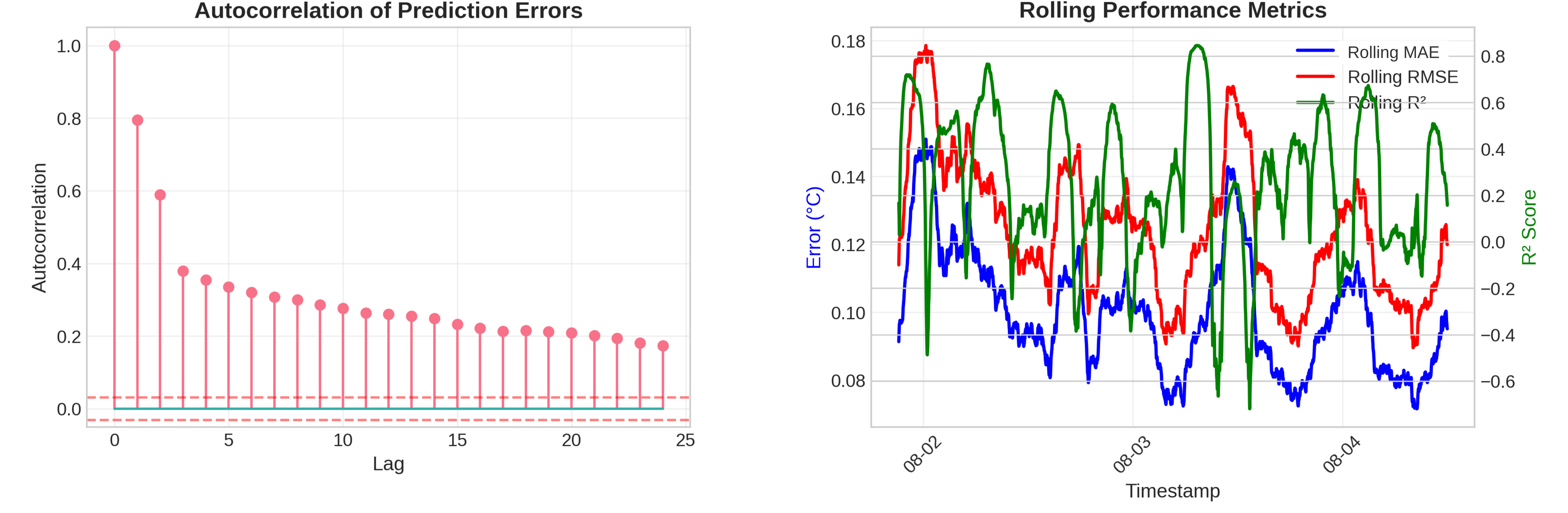}
    \caption{Extended Performance Analysis. (Left) Autocorrelation of prediction errors shows no significant patterns beyond lag 0, indicating that the model has captured the temporal dynamics of the data without leaving systematic patterns in the residuals. (\x{Right}) Rolling performance metrics (\x{MAE}, RMSE, R²) remain stable over the evaluation period, showing model robustness across different time segments.}
    \label{fig:error_analysis}
\end{figure*}

\begin{figure*}[ht!]
    \centering
    \includegraphics[scale=0.1]{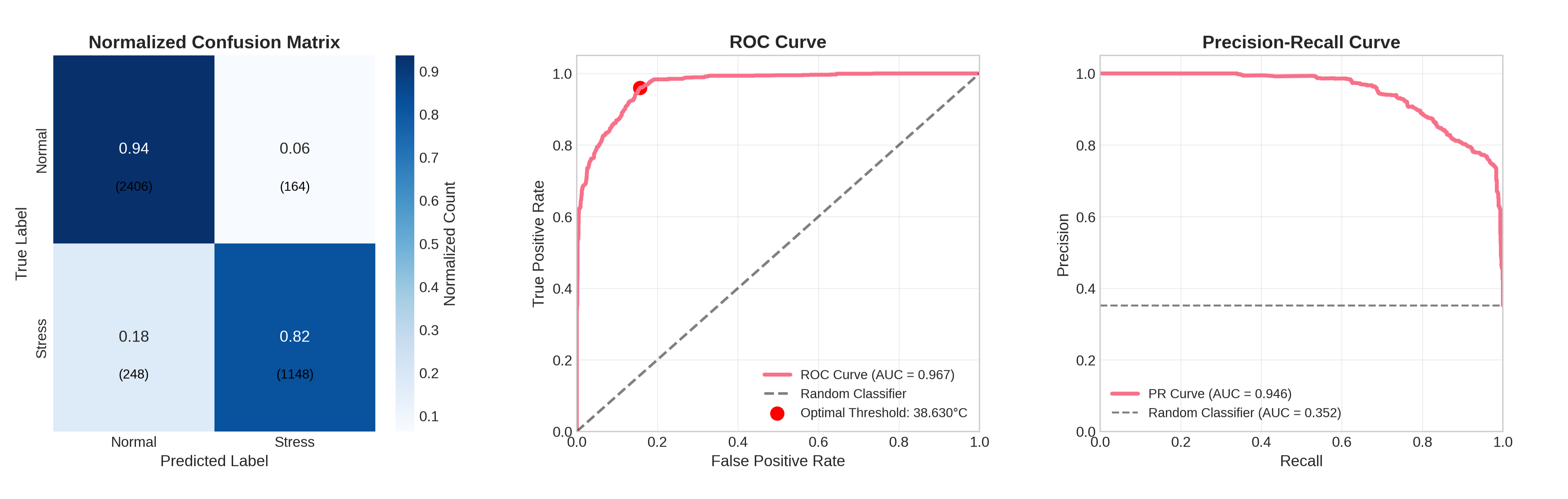}
    \caption{Classification Performance for Heat Stress Prediction. (Left) Normalized Confusion Matrix showing high accuracy for both Normal (0.94) and Stress (0.82) classes. (Middle) ROC Curve with AUC of 0.967, demonstrating good discriminative ability. The optimal threshold is identified at 38.630°C. (Right) Precision-Recall Curve with AUC of 0.946, indicating strong performance across different classification thresholds. The model achieves an accuracy of 0.8938, precision of 0.8411, recall of 0.8453, and F1-score of 0.8425.}
    \label{fig:classification}
\end{figure*}

Figure~\ref{fig:regression_analysis} provides a comprehensive view of the model's regression performance. The prediction vs. true values scatter plot shows a strong linear relationship with points clustered closely around the perfect prediction line. The error distribution histogram reveals that the majority of prediction errors are small and normally distributed around zero, with the model achieving a mean absolute error of only 0.114°C. The residual plot confirms the absence of systematic bias, as the trend line remains flat and residuals are randomly distributed. The \x{bottom} subplot demonstrates the model's ability to quantify prediction uncertainty, with the true values consistently falling within the predicted confidence intervals.

Further analysis of the model's residuals and rolling performance metrics (Figure~\ref{fig:error_analysis}) confirms its robustness. The prediction errors show no significant autocorrelation, indicating that the model has captured the temporal dynamics of the data without leaving systematic patterns in the residuals. The rolling $R^2$, MAE, and RMSE remain stable over time, demonstrating that the model's performance is consistent and not prone to degradation over different periods.

\subsection{Classification Performance Analysis}

The derived classification model for heat stress events shows outstanding discriminative power. As depicted in Figure~\ref{fig:classification}, the model achieves an {AUC of 0.967} on the ROC curve and {0.946} on the Precision-Recall curve. The normalized confusion matrix reveals high classification accuracy for both Normal (0.94) and Stress (0.82) classes, with the optimal classification threshold identified at 38.630°C. \x{This $38.630^{\circ}$C corresponds to the ROC-optimal \emph{operating point} of the classifier, obtained as the Youden-$J$ cut on the predicted-CBT axis. It is distinct from the $38.8^{\circ}$C clinical threshold that defines the ground-truth events: the operating point is a tunable decision boundary, whereas the clinical threshold is a fixed physiological criterion. Placing the operating point slightly below the clinical threshold provides earlier warning while leaving the definition of a heat-stress event unchanged.} The high precision (0.8411) and recall (0.8453) indicate its effectiveness for real-world deployment, capable of minimizing both costly false alarms and dangerous missed events.




\section{Discussion}
\label{sec:discussion}
This study presents a physics-informed, uncertainty-aware multimodal forecasting framework substantially advancing CBT prediction in dairy cattle. By embedding a physiologically grounded DT within an expert-based stacked ensemble, our approach addresses multiple methodological gaps identified in the literature: lack of physiological grounding \cite{11, 13}, limited long-horizon forecasting \cite{2, 4, 5}, absence of uncertainty quantification \cite{2,4,5,8,12,13,oliveira2025heat}, heuristic fusion strategies \cite{2, 3, 5}, and a lack of personalization. These limitations motivate our design choices in (Section \ref{sec:methodology}), where thermodynamic constraints, reliability-aware fusion, and calibrated predictive intervals are integrated into a single causal forecasting pipeline. \cite{2, 3, 4, 5, 6, 7, 8, 11, 13, oliveira2025heat}.

A key strength of our framework lies in its foundation on fundamental scientific principles. The DT's physiology model is derived directly from the First Law of Thermodynamics, grounding the core body temperature predictions in the universal conservation of energy. The framework captures the actual mechanistic dynamics of thermoregulation in cattle by explicitly modeling the balance between metabolic heat production, activity-induced heat generation, environmental heat absorption, and heat dissipation. This principled approach improves interpretability compared to purely data-centric models. It provides a robust inductive bias that guides ML components, allowing them to focus on residual, individual-specific patterns without violating known biological or physical laws.

Our results demonstrate a marked performance improvement over existing methods. The proposed model achieved an overall cross-validated $R^2$ of 0.7835, a considerable gain relative to traditional baselines and prior work. For instance, Li et al. \cite{8} forecasted CBT using data-centric models and achieved a maximum $R^2$ of 0.54 with a GWO--XGBoost ensemble; in comparison, our approach represents an absolute improvement of approximately 0.24 in $R^2$ (0.7835 vs. 0.54) under our cross-validated evaluation protocol. Moreover, constraining dynamics through conservation-based priors reduces sensitivity to spurious correlations in noisy multimodal streams, improving stability during partial observability (e.g., intermittent sensor dropout) compared to unconstrained statistical surrogates. This performance leap stems from two key innovations. First, the physics-informed DT, which integrates a first-principles ODE model, Gaussian Process residual modeling, and Kalman Filter synchronization, provides physiologically meaningful, individualized features that statistical surrogates lack. Ablation studies confirmed that incorporating DT features increased R² by 13.06\% and reduced MAE by 17.22\% compared to models without DT features. This gain is consistent with the view that DT features act as physiologically meaningful state estimates, filtering noisy observations into dynamics-aligned representations that are easier for downstream learners to extrapolate over longer horizons. This highlights the critical role of physiological priors in improving forecast fidelity \cite{abbott2018vivo,roseler1997development,felini2024assessing}.

Furthermore, replacing the LightGBM backbone with deep temporal models such as Transformer or LSTM-based hybrids did not yield performance improvements. This outcome follows from a fundamental architectural property of the framework: the physics-informed DT already performs the temporal representation learning that deep sequence models typically learn through an end-to-end process. The ODE captures thermal trends, the Kalman filter tracks rate-of-change, the Markov chain encodes behavioral transitions, and multi-scale rolling statistics (Section~\ref{subsec:features}) encode short- to long-range temporal patterns explicitly. The residual learning task is therefore predominantly cross-modal and nonlinear-tabular rather than sequential, in which gradient-boosted trees consistently match or outperform deep architectures on structured data~\cite{grinsztajn2022tree,shwartz2022tabular}. This is empirically confirmed by our baselines (Table~\ref{tab:baseline_comparison}), where both LSTM ($R^2 = 0.367$) and Temporal Fusion Transformer ($R^2 = 0.316$) underperformed even standard XGBoost ($R^2 = 0.590$) despite receiving identical engineered features. LightGBM further offers native handling of missing values and mixed feature types (continuous sensor readings, categorical behavioral states, DT-derived probabilities), transparent feature importance, and sub-millisecond inference without GPU acceleration, all of which align with the deployment and interpretability requirements of precision livestock farming. Extending the framework with hybrid Transformer--DT architectures remains a promising direction as larger multi-farm cohorts become available.

Second, our modality-specific expert weighted fusion mechanism systematically weights data streams according to their validation reliability, moving beyond the heuristic concatenation used in previous multimodal studies \cite{2,3,4,5, 13}. Prior work on multimodal fusion for classification tasks, such as Dhaliwal et al. \cite{2} and Tong et al. \cite{3}, demonstrated high within-dataset accuracy but had limitations: Dhaliwal et al. \cite{2} relied on physiologically meaningful cues without a formal adaptive reliability mechanism, while Tong et al. incorporated adaptive scoring but did not utilize detailed physiological indicators. Our results show that adaptive fusion is essential for robust performance: single-modality models achieved at most $R^2$ = 0.3786, whereas the complete fusion model reached R² = 0.7835, underscoring the complementary value of combining DT-derived, behavioral, environmental, and physiological data streams.

Uncertainty quantification, which has been largely absent from prior CBT forecasting efforts \cite{13}, is another distinctive contribution of this study. By applying bootstrap-based predictive intervals and validating them with the PICP, we achieved PICP of 92.38\%. This calibration level is significant for deployment scenarios, where decision thresholds for cooling interventions must balance false alarms and missed detections. Unlike previous works that reported only point estimates \cite{8, 11, 13}, our framework provides well-calibrated predictive uncertainty, addressing limitations highlighted by Woodward et al. \cite{13}.

\x{The framework operates as an autoregressive multi-horizon forecaster: it consumes the full multimodal history up to time $t$, including past CBT, and predicts CBT at $t+120$~min. Its predictive value therefore derives from the physics-informed multimodal fusion and the calibrated uncertainty, and its deployment depends on the availability of the CBT channel. Where continuous intravaginal logging is available, this channel is used directly. Where CBT is available only intermittently or is absent, the thermal state is instead propagated by the ODE, the behavioral Markov model, and the environmental inputs between readings, while the remaining modalities continue to contribute. The per-modality ablation in Table~\ref{tab:ablation_modality}, in which the non-CBT groups retain weaker but non-negligible predictive signal and the strongest single group reaches $R^2$ equivalent to $0.3786$, indicates the degradation expected as the CBT channel becomes sparse. A fully sensor-only configuration, obtained by removing the physiological-CBT feature group together with the Kalman CBT update, is a natural extension left to future work.}

Our framework also moves beyond population-level averages to individualized modeling. The DT’s Gaussian Process residual component captures cow-specific deviations, while GroupKFold cross-validation ensures animal generalization. However, while GroupKFold mitigates within-animal leakage, it does not fully guarantee external validity across farms, climates, or sensor hardware, motivating broader evaluation. This contrasts with the small, within-herd datasets commonly used in prior multimodal classification studies \cite{2, 4, 5}, often involving fewer than a dozen animals.

While our framework demonstrates strong predictive performance and robust uncertainty quantification, one consideration regarding the benchmark dataset remains. To build upon these advancements, future validation should include multi-farm and multi-breed cohorts using leave-one-farm-out cross-validation. This should be coupled with farm-specific recalibration of DT parameters (e.g., $\beta_{tolerance}$ priors) and domain-robust fusion weighting to properly quantify the model's transferability under heterogeneous management practices and climatic conditions. Despite these limitations, the results support the theoretical premise that combining physiological models with data-centric learning improves accuracy and interpretability. The proposed system enables real-time, individualized forecasting of heat stress two hours in advance with high predictive confidence, supporting timely interventions such as targeted cooling or shade provision. This early-warning capability is evaluated explicitly at a 2-hour prediction horizon under causal inputs and GroupKFold splits, supporting deployment-oriented decision support rather than retrospective detection. This represents an advancement over prior approaches that rely primarily on environmental proxies \cite{11, 13} or use daily-aggregated cow-level data \cite{12}, both lacking the temporal granularity and uncertainty-aware modeling required for precision livestock farming.

Regarding scalability and deployment feasibility, the proposed framework integrates multiple modeling components, yet each remains computationally efficient at inference time. The ODE-based thermoregulation model performs a single first-order Euler integration per time step, the Markov behavioral model requires only a $4 \times 4$ matrix--vector multiplication, and the Kalman filter operates on a compact three-dimensional state vector ($T_{\text{core}}$, $\dot{T}_{\text{core}}$, $A_{\text{level}}$), all of which execute in constant time per update. The Gaussian Process operates on bounded per-cow residual windows rather than the full dataset, keeping its cubic cost tractable. On the ML side, LightGBM inference traverses pre-trained decision trees with sub-millisecond latency per sample, and the three-stage stacked ensemble adds only a small constant factor over a single model evaluation. Crucially, all computationally intensive operations, including Optuna hyperparameter search, bootstrap resampling, and GroupKFold cross-validation, are performed offline during model training and are not part of the real-time inference path. Since the pipeline operates at a 1-minute temporal resolution, the combined per-step inference budget is well within the capacity of commodity edge hardware. Consequently, the framework is amenable to deployment on farm-level computing infrastructure , supporting practical adoption in precision livestock farming environments.

\section{Conclusion}
\label{sec:conclusion}
This study introduces a novel physics-informed, uncertainty-aware, multimodal forecasting framework that significantly improves the long-horizon prediction of CBT in dairy cattle. By integrating a physiologically grounded DT with adaptive expert-based data fusion and rigorous uncertainty quantification, the proposed approach achieves SOTA performance with an $R^2$ of 0.783, well-calibrated predictive intervals having a PICP of 92.38\%, and substantial heat stress detection capability with an F1 score of 0.8425 and an area under the curve of 0.967. Additionally, we include theoretical contributions by demonstrating the value of embedding physiological priors in ML pipelines and weighting modalities according to predictive reliability. 

This work contributes to both theory and practice within precision livestock farming. The framework addresses key limitations of prior approaches by providing physiologically interpretable features, individualized predictions, calibrated uncertainty estimates, and robust multimodal data fusion. Unlike traditional weather-only proxies or purely statistical surrogates, the proposed model captures the dynamic interplay between environmental, behavioral, and physiological signals, allowing for the early detection of heat stress with high confidence. This enhances decision-making capabilities, demonstrating a clear pathway from advanced computational modeling to real-world application.

Future work will extend the framework to multi-farm settings, refine physiological parameter calibration through field measurements, and integrate adaptive decision policies comprising the uncertainty estimates for dynamic intervention strategies. In addition, this work presents the foundation for physiologically grounded, reliable, and generalizable DTs for precision livestock management.

\section*{Declarations}

\noindent
\textbf{Conflict of interests:} On behalf of all authors, the corresponding author states that there is no conflict of interest.

\noindent
\textbf{Funding:} No external funding is available for this research.

\noindent
\textbf{Data availability statement:} The publicly available MmCows: A Multimodal Dataset for Dairy Cattle Monitoring \cite{vu2024mmcows} was utilized in this study.

\noindent
\textbf{Ethics approval and consent to participate:} Not applicable.

\noindent
\textbf{Informed consents:} Not applicable.

\textbf{CRediT Author Statement}

\textbf{Riasad Alvi}: Conceptualization, Methodology, Software, Formal analysis, Investigation, Data curation, Writing -- Original Draft, Visualization.
 \textbf{Mohaimenul Azam Khan Raiaan}: Conceptualization, Methodology, Supervision, Validation, Formal analysis, Writing -- Original Draft, Writing -- Review \& Editing, Project administration. \textbf{Sadia Sultana Chowa}: Writing -- Original Draft, Writing -- Review \& Editing, Visualization. \textbf{Arefin Ittesafun Abian}: Supervision, Validation, Visualization, Writing -- Review \& Editing. \textbf{Reem E Mohamed}: Validation, Visualization, Writing -- Review \& Editing. \textbf{Md Rafiqul Islam}: Validation, Visualization, Writing -- Review \& Editing. \textbf{Yakub Sebastian}: Validation, Visualization, Writing -- Review \& Editing. \textbf{Sheikh Izzal Azid}: Validation, Visualization, Writing -- Review \& Editing. \textbf{Sami Azam}: Supervision, Validation, Visualization, Writing -- Review \& Editing.








	
\end{document}